\documentclass{article}
\pdfoutput=1

\usepackage[utf8]{inputenc}
\usepackage{arxiv}
\usepackage{hyperref}
\usepackage{url}
\usepackage{graphicx}
\usepackage{comment}
\usepackage{tabulary}
\usepackage{subcaption}
\usepackage{listings}
\usepackage{tabulary}
\usepackage{colortbl}
\usepackage{array}
\usepackage{amsmath}
\usepackage{float}
\usepackage{doi}

\renewcommand{\headeright}{Enhanced Version}
\renewcommand{\undertitle}{Enhanced Version}
\newenvironment{notice}
{
  \cleardoublepage
  \centerline
  {\large \bfseries \scshape Notice}
}
{
  \cleardoublepage
  \thispagestyle{empty}
}
\hypersetup{pdfauthor={Luis Rei},pdftitle={Multimodal Metadata Assignment for Cultural Heritage Artifacts}}

\begin{document}
\title{Multimodal Metadata Assignment for Cultural Heritage Artifacts}

\author{
\href{https://orcid.org/0000-0001-8587-1638}{\includegraphics[scale=0.06]{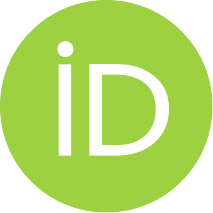}\hspace{1mm}Luis Rei} \\
	Department for Artificial Intelligence\\
	Jošef Stefan Institute\\
	Ljubljana, Slovenia \\
	\texttt{luis.rei@ijs.si} \\
\And
\href{https://orcid.org/0000-0002-0360-6505}{\includegraphics[scale=0.06]{orcid.pdf}\hspace{1mm}Dunja Mladenić} \\
	Department for Artificial Intelligence\\
	Jošef Stefan Institute\\
	Ljubljana, Slovenia \\
	\texttt{dunja.mladenic@ijs.si} \\
\And
\href{https://orcid.org/0000-0002-1293-6039}{\includegraphics[scale=0.06]{orcid.pdf}\hspace{1mm}Mareike Dorozynski} \\
	Institute of Photogrammetry and GeoInformation\\
	Leibniz University Hannover\\
	Hannover, Germany \\
	\texttt{dorozynski@ipi.uni-hannover.de} \\
\And
\href{https://orcid.org/0000-0003-1942-8210}{\includegraphics[scale=0.06]{orcid.pdf}\hspace{1mm}Franz Rottensteiner} \\
	Institute of Photogrammetry and GeoInformation\\
	Leibniz University Hannover\\
	Hannover, Germany \\
	\texttt{rottensteiner@ipi.uni-hannover.de} \\
\And
\href{https://orcid.org/0000-0002-9879-2004}{\includegraphics[scale=0.06]{orcid.pdf}\hspace{1mm}Thomas Schleider} \\
	EURECOM \\
	Biot, France \\
	\texttt{thomas.schleider@eurecom.fr} \\
\And
\href{https://orcid.org/0000-0003-0457-1436}{\includegraphics[scale=0.06]{orcid.pdf}\hspace{1mm}Raphaël Troncy} \\
	EURECOM \\
	Biot, France \\
	\texttt{raphael.troncy@eurecom.fr} \\
\And
\href{https://orcid.org/0000-0002-5680-1777}{\includegraphics[scale=0.06]{orcid.pdf}\hspace{1mm}Jorge Sebastián Lozano} \\
	Departamento de Historia del Arte \\
    Universitat de València \\
	Valencia, Spain \\
	\texttt{jorge.sebastian@uv.es} \\
\And
\href{https://orcid.org/0000-0003-0680-9364}{\includegraphics[scale=0.06]{orcid.pdf}\hspace{1mm}Mar Gaitán Salvatella} \\
	Departamento de Historia del Arte \\
    Universitat de València \\
	Valencia, Spain \\
	\texttt{m.gaisal@uv.es} \\
}

\maketitle
\thispagestyle{empty}

\begin{abstract}
We develop a multimodal classifier for the cultural heritage domain using a late fusion approach and introduce a novel dataset. The three modalities are Image, Text, and Tabular data. We based the image classifier on a ResNet convolutional neural network architecture and the text classifier on a multilingual transformer architecture (XML-Roberta). Both are trained as multitask classifiers and use the focal loss to handle class imbalance. Tabular data and late fusion are handled by Gradient Tree Boosting. We also show how we leveraged specific data models and taxonomy in a Knowledge Graph to create the dataset and to store classification results. All individual classifiers accurately predict missing properties
in the digitized silk artifacts, with the multimodal approach providing the best results.
\end{abstract}


\begin{notice}
\raggedright
Dear reader,

This document is a modified version of the Preprint Version 1 \doi{10.21203/rs.3.rs-1708875/v1}, which was subsequently published as \doi{10.1007/s00530-022-01025-2}. Please consider citing the published version:

\begin{quote}
   Rei, L., Mladenic, D., Dorozynski, M. et al. Multimodal metadata assignment for cultural heritage artifacts. Multimedia Systems 29, 847–869 (2023). \url{https://doi.org/10.1007/s00530-022-01025-2}
\end{quote}

Significant enhancements in this version include a new comparison of strategies for addressing class imbalance detailed in Section~\ref{sec:exp}, alongside their description in Section~\ref{sec:balance} and discussion in Section~\ref{sec:conclusions}. We also added the confusion matrices in Section~\ref{sec:exp_analysis}. Additional minor improvements have been made throughout the article.

This version was intended for publication; however, due to an oversight, it was not published. Efforts to have the journal replace the published version with this one were unsuccessful. Believing the additional content in this version to be of value, I am making it available publicly.

Thank you for your interest in this work.\\
Luis Rei
\end{notice}

\section{Introduction}
\label{sec:introduction}

\subsection{Motivation}
Some cultural heritage domains deal with knowledge that is not broadly known by the public, but only by domain experts. Despite many objects having been digitized, even those experts still struggle to find what they are looking for in online catalogs. Thus, they are forced to return to the cumbersome manual consultation of published catalogs or even card files. If such a situation arises for experts, the broader public is still more removed from access to that information. The European production of silk fabrics is an example of one such domain. It is witness to an essential field of European and global history, linked to world trade routes, the production of luxury goods of enormous symbolic importance, technological developments, and the very advent of the Industrial Revolution. However, the material vulnerability of these objects and the institutional fragility of many local heritage organizations have rendered them relatively hidden from the public. As regards the information about that heritage, many descriptions, and images of objects exist within in-house databases that are only available as local files. In other cases, those records are uploaded by many museums across the globe, in siloed repositories and heterogeneous, often incompatible formats. A few of them provide public access to the images and metadata of their silk objects through APIs, and many more through their websites, but harmonization and integration efforts have been very scarce. Therefore, it is very hard for general audiences, historical experts, and industry (e.g., fashion designers) to access this knowledge.

One solution that can help alleviate this problem for a cultural heritage domain such as European silk fabrics is an Exploratory search engine, which helps users to explore a topic of interest~\cite{Palagi:PhD2018}. They enable serendipitous discovery of items, and they are especially appropriate when these items come with rich structured metadata. ADASilk~\footnote{https://ada.silknow.org/}, named after Ada Lovelace, is an exploratory search engine, based on a knowledge graph (KG). ADASilk enables both domain experts and users not necessarily familiar with this topic to search and browse silk fabric objects. Thus, not only historians or scholars but also designers or simply fans of fashion can access such a significant and little-known part of our heritage.

Some records in cultural archives have what can be considered essential information, like the production year or the weaving techniques, semantically annotated, others include it only in rich textual descriptions, and for some objects, it is not available in any form. These missing metadata can be considered as gaps that could be potentially filled in. Thanks to the progress in natural language processing, information extraction, and image processing, there are now techniques that can help to address such problems. Digitization of culturally significant assets is a time-consuming process that requires experts and funding. This often forces a cultural institution to make a trade-off between the number of objects digitized and the effort per object. Less effort per object often implies a smaller number of details captured, less strict guidelines, and sometimes mistakes. Nevertheless, this area could benefit from automated aids for collection caretakers, which often must catalog similar or identical objects scattered across the world. Obtaining predictions or suggestions for their description and possible matching pieces would be a great help for that task, taking also into account the many objects still waiting to be properly cataloged.

\begin{figure}[hbt]
    \centering
     \begin{subfigure}[b]{0.3\textwidth}
         \centering
         \includegraphics[width=\textwidth]{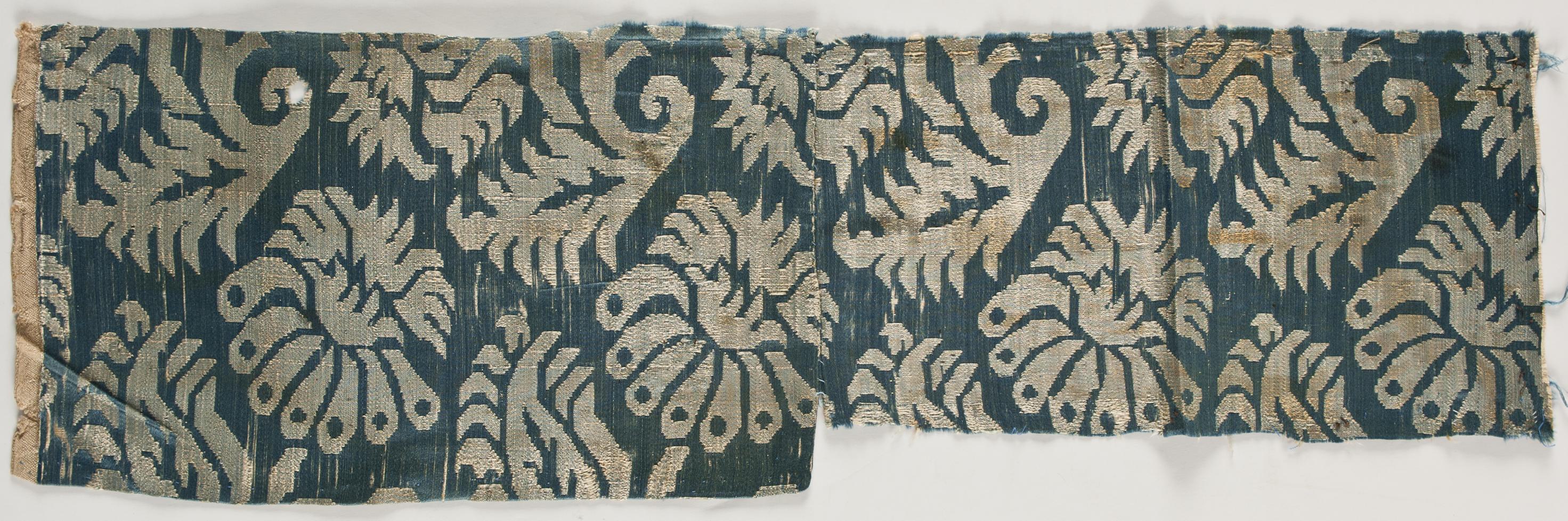}
         \caption{XVII/Animal fibre}
         \label{fig:img_examples:1}
     \end{subfigure}
    \hfill 
     \begin{subfigure}[b]{0.15\textwidth}
         \centering
         \includegraphics[width=\textwidth]{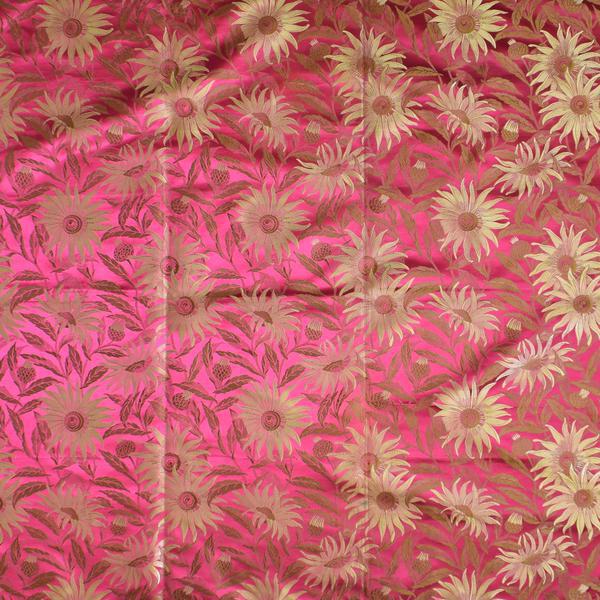}
         \caption{GB/XIX}
         \label{fig:img_examples:2}
     \end{subfigure}
    \hfill
     \begin{subfigure}[b]{0.18\textwidth}
        \centering
         \includegraphics[width=\textwidth]{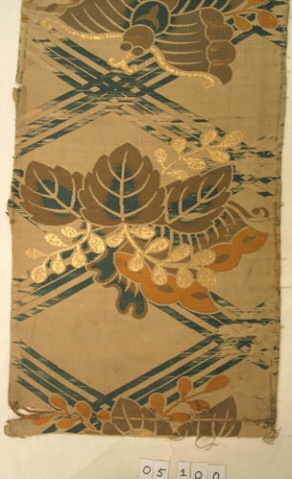}
         \caption{JP/Animal fibre}
         \label{fig:img_examples:3}
     \end{subfigure}   
    \hfill
      \begin{subfigure}[b]{0.26\textwidth}
         \centering
         \includegraphics[width=\textwidth]{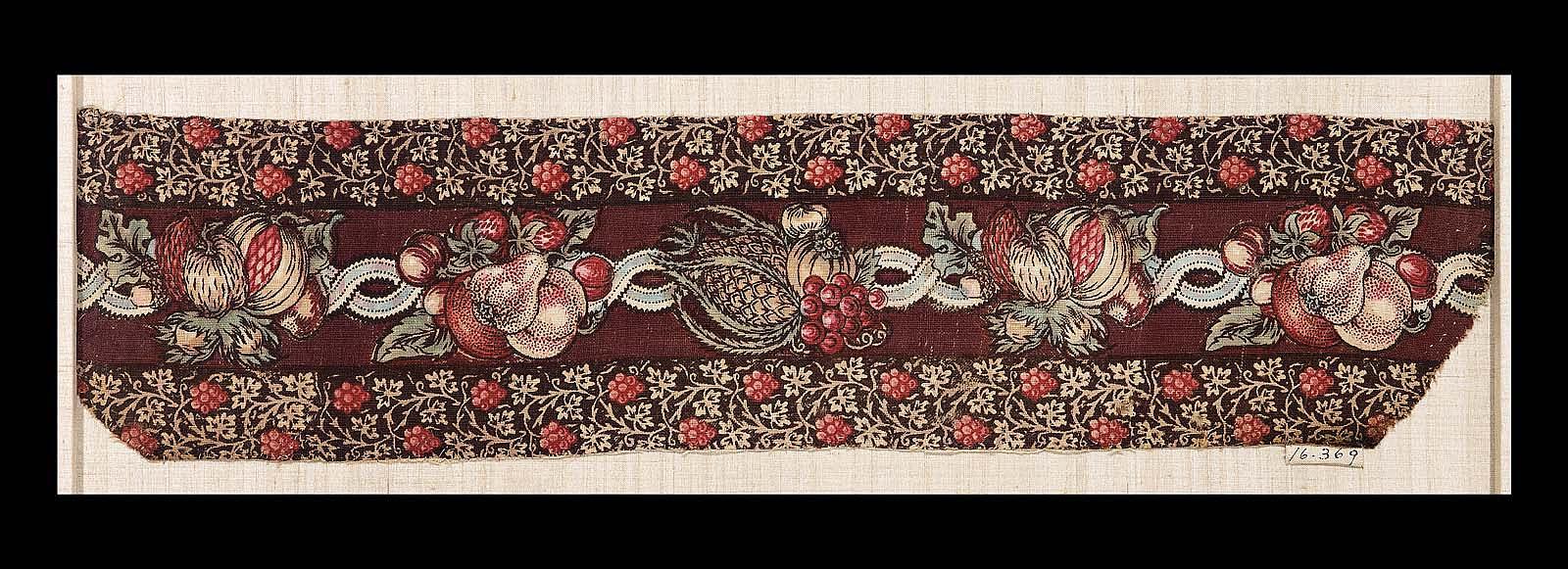}
         \caption{GB/XVII}
         \label{fig:img_examples:4}
     \end{subfigure}

        \caption{Images in our dataset with labels}
        \label{fig:img_examples}
\end{figure}

This article presents methods that enable further annotation of these museum objects through a multimodal classification approach that trains models to predict missing metadata from images, text descriptions, and other (available) metadata. The outcome is then further used to enrich an underlying knowledge graph that feeds the ADASilk exploratory search engine. Domain experts can easily assess the quality of the automatically generated annotations through rich visualization and connections between the items. Figure \ref{fig:img_examples} shows examples of the kinds of images we are discussing. While Table~\ref{tab:text_examples} shows some examples of text descriptions from various museums. Both show respectively their associated metadata.

\tiny
\begin{table}[hbt]
\centering
\caption{Texts in our dataset with labels}
\label{tab:text_examples}
\begin{tabulary}{\linewidth}{L}
    \hline\noalign{\smallskip}
    White and silver striped fabric with supplementary weft of flat silver strips whose floats form vertical stripes with leaves at intervals. White floats of the weft form outlines for serpentine floral sprays spread over the striped areas. \\
    \hfill FR/Animal fibre \\
    \noalign{\medskip}
    Furnishing fabric, woven, British, c. 1895, Alexander Morton \& Co., red/brown plain silk weave \\
    \hfill GB/XIX  \\
    \noalign{\medskip}
    Dibujo Palma en color azul grisáceo Urdimbre: Trama: 36 pasadas Rapport: 65 cm ancho y 104 cm alto (incompleto) \\
    \hfill ES/XX/Damask/Animal fibre \\
    \noalign{\medskip}
    Alentours de : Jean-Baptiste Blain de Fontenay \& Claude Audran. Carton 1751-1757 Tissage achevé 1758 ancienne tenture n° 225 - 12 pièces complémentaires. Atelier d'AUDRAN. Signature : AUDRAN G. (fleur de lys) 1758 5ème alentour dit de Marly, fond jaune d'ornements mosaïque. Fonds des Gobelins  \\
    \hfill XVIII/Animal fibre \\
    \noalign{\smallskip}\hline
\end{tabulary} 
\end{table}
\normalsize

\subsection{Hypothesis}
\label{sec:introduction-hypothesis}
 
Our first hypothesis is that we can fairly accurately predict a set of domain-relevant properties of cultural heritage objects (silk fabrics) from images and text descriptions.
Our second hypothesis is that a multimodal approach involving both images, text descriptions, and properties other than those to be predicted will produce better results than any method relying on a single modality. 
In this context, the term "better" refers to both, the quality of the results and the number of objects for which this information is inferred.
That is, we expect the multimodal approach to result in more correct predictions and in predictions for a larger number of objects than the single modality methods. 
These hypotheses will be evaluated in the context of digitized metadata of silk fabric artifacts with data originating in multiple museums.

\subsection{Contributions}
The main scientific contributions of this paper are related to our research hypotheses. We introduce a multimodal machine learning approach, adapted to the cultural heritage domain, for predicting the properties of digitized artifacts. 
We perform an in-depth analysis of the performance of our classification models, i.e., models based on individual modalities and the multimodal classifier. 
Additionally, we introduce a novel dataset\footnote{\url{https://zenodo.org/record/6590957}} to the cultural heritage and multimodal analysis domains that includes data for four different tasks and three different modalities. It consists of harmonized text and image data from heterogeneous, multilingual sources that went through different stages of preprocessing, cleaning, and enrichment like domain expert-guided entity linking and grouping. 

Finally, we show how our metadata predictions can be properly represented within our data model with appropriate classes and properties. This includes using information such as their time stamp, the algorithm used, and confidence. The actual predictions can consequently be integrated into the existing Knowledge Graph.

\subsection{Challenges}
The challenges faced in this work can be split broadly into those pertaining to the creation of the dataset and those related to automated annotation. 
The latter can be further categorized according to the modality that is used for predicting the properties of the objects.

\subsubsection{Data and Labels}
The data used in this work belongs to the cultural heritage domain. More specifically, it is related to silk textiles produced in Europe, primarily in the period between the 15th and the 20th centuries. In the domain of cultural heritage, we cannot expect all class labels to be equally common or equally correlated. For example, in some locations, more silk fabric objects were produced than in others. Similarly, we know that the production of silk fabric objects in a given location likely intensified after a certain point in time and possibly subsided after a certain date. We also know that catalogs are curated by humans and often have strong thematic biases. For example, certain museums focus almost exclusively on objects created within one location.
The data we use in this work was aggregated from 12 different museum or collection websites. Each museum may have different standards for how it collected the underlying objects and how it digitized the information related to these objects. Importantly, this gives each museum its own standards for how to write text descriptions, how to create images, and how to annotate properties. Regarding these properties of digitized artifacts, accurately representing them requires adequate data modeling capabilities and considerable domain expert collaboration. This collaboration is also important in creating a dataset for machine learning. Labels need to be mapped from annotations made in different languages and grouped into domain-relevant classes. Due to the partially automated nature required to create the dataset, challenges arise that are common in such processes: label text requires normalization such as correcting typos, unifying the styles of dates, and matching different locations to specific countries. Errors made in this process can often be systematic, for example, a failure to link a specific value of a property due to the way it is written will likely result in that value not being present in all records originating in that museum.

\subsubsection{Image Classification.} 
In the context of this paper, the classification of images aims to predict the abstract properties of the silk fabrics depicted in the images.
Whereas it may be relatively straightforward to learn to classify the material of a depicted piece of fabric, the prediction of semantic information such as the production place of the fabric, the period of time in which the fabric was produced, or the technique used to manufacture the fabric is assumed to be much more challenging. Furthermore, it is assumed that there are interdependencies between these properties of silk fabrics, e.g., a certain production technique may have been more common during a certain period of time.
This is why multitask learning is investigated for image classification. However, standard multitask classification frameworks require one reference label for every task to be learned during training for every training sample; 
The challenge we have to face is that in real-world data, as they were collected for the dataset presented in this paper, there may be many training samples for which annotations are unavailable for some target variables to be predicted. Accordingly, this fact must be taken into account in the training of a multitask classifier.
Additionally, the available number of class labels constituting the class distribution of a variable is often imbalanced for real-world datasets.
This constitutes a further challenge to supervised learning, which is addressed by utilizing a suitable training strategy for the image classification method.

\subsubsection{Text Classification}
Supervised approaches are often challenging to perform with data from the cultural heritage domain for several reasons. Text descriptions are not present for the majority of objects in an archive. Many of the text descriptions that are available, in most museums, tend to be short sentences, almost title-like. In specific domains, such as the cultural heritage of silk production, many of the terms used in the text are very domain-specific. Each museum has its own standard of how and what to write in a text description. Some may focus on the history of the objects and write very grammatical paragraph-length descriptions meant to be read by the public. Others may focus on the properties of the object and write a single enumerating sentence. Others, still, may focus solely on the depictions or visual patterns of an object. Finally, museums are spread geographically, and thus we can expect to deal with multiple languages, making our problem multilingual and cross-lingual. To summarize, we end up with a small collection of domain-specific texts, written in different languages, with different content both semantically and syntactically, and wildly varying lengths. These texts are then associated with labels, based on the provided properties of the object. As already discussed, these labels are not all equally likely or correlated, and many of these accidental regularities are likely to interact with the language and the particularities of the text style of the museum.

\subsubsection{Multimodal Classification}
One of the challenges in this work is that we want to integrate predictions made from images and text. Most work done in the literature is exclusive to depictions or the type of objects: the image shows a scene or object and the text describes it. In our case, there may be no scene depicted in an object, and we do not consider describing the object beyond certain properties. For example, if we have a fabric that shows a certain pattern, describing the visual shapes of the pattern (e.g., triangles) is not a goal. Rather, we need to deduce the image properties of how, when, where, and with what the object was made. Similarly, with text descriptions, there may be a good amount of words that describe visual patterns, scenes depicted, and historical facts associated with the object, but the goal is, again, to determine those same intrinsic properties of the object's making. 
Another uncommon challenge is the reduced variable overlap between images and text descriptions. Not only is our work subject to a comparatively small dataset, restricted by historical reality and difficulties of data collection, but we must also deal with the fact that for most archives of culturally relevant objects, many objects that have been photographed have no corresponding textual description. In fact, we'll see that less than half of all objects have both these modalities. 
Another challenge, uncommon outside of retrieval scenarios, is that we can have multiple images with different angles and focus per each individual object while it makes no sense to talk about multiple text descriptions per object. Yet another challenge we need to deal with, common to many real-world applications but not to research datasets, is that we do not have all properties for all objects. For example, for a given object, we might know what material and techniques were used but not when or where it was made. Finally, our dataset, although drawn from several museums, contains under 30k objects and approximately 11k text descriptions. This effectively makes it small compared to general datasets, but not uncommonly so for a dataset in the cultural heritage domain.

\section{Related Work}

\subsection{Cultural Heritage Domain}
Since the development of the web, many Cultural Heritage (CH) organizations have provided metadata on their items through search engines, APIs, or aggregators. Unfortunately, there has been little unity in the data formats, which makes data integration a complex task. One solution to this problem in the case of museum data is the use of Semantic Web technology, and more specifically the development of Knowledge Graphs based on ontologies that follow the open CIDOC Conceptual Reference Model (CRM). CIDOC-CRM was developed for this purpose, i.e., to facilitate inter-museum data integration. It provides many relevant classes and properties to represent domain-specific CH objects and is easily extendable. It is the outcome of more than 20 years of development by ICOM's International Committee for Documentation (CIDOC)~\cite{Doerr2005TheCC}. CIDOC-CRM can, however, only be considered a starting point for ontologies that deal with museum data, such as in our case. The fact that it is not difficult to extend it makes it, however, easy to add more domain-specific classes and properties as necessary in projects such as ours. 
 
 There are more and more efforts of different CH organizations to adopt Semantic Web technologies and build knowledge graphs: CultureSampo is the result of integrating heterogeneous cultural content~\cite{cultureSampo}. The challenges consisted amongst others of converting legacy data into linked data and making it heterogeneous. Getty ULAN was used as structured vocabulary to find connections between two referenced persons, for example. One similar example is ArchOnto~\cite{archonto}, which specifically addresses the challenges of CH data from and for national archives. Both can be an inspiration for our work, but given how fine-grained the vocabularies in Cultural Heritage domains can be, it is still necessary to deal with language and domain-specific vocabulary differently in each case.
 
 The training data used for our experiments is fully extracted from the SILKNOW Knowledge Graph that relies on classes and properties defined by CIDOC-CRM and its direct extensions CRMsci (Scientific Observation Model) and CRMdig (Model for provenance metadata). All our data is therefore part of the specific CH domain of "silk fabrics" and accordingly semantically annotated and enriched, for example through linking and normalization of properties, such as materials used and weaving techniques.

\subsection{Knowledge Graphs and Culture AI}
Knowledge graphs allow the representation of multi-source information about many entities and their relationships to each other. The data stored in a Knowledge graph can then be used for many tasks, especially when structured knowledge of a specific domain is relevant, e.g., the development of product designs~\cite{KG-aided2020}. Other common domain-specific fields are Medicine, Cybersecurity, and Finance. Knowledge graphs are also frequently used to aid product development and research for language-based tasks such as question-answering systems, recommender systems, and information retrieval~\cite{Zou_2020, DBLP:journals/corr/abs-2004-00387}. A knowledge graph can also help with textual metadata-aided visual pattern extraction and recognition~\cite{castellano2021deep}. Lastly, as we deal not only with images but also textual metadata, the SemArt project can be considered related: it is a multi-modal dataset for semantic art understanding. Unlike in this study, they did, however, not work towards metadata completion, but focused only on retrieval~\cite{garcia2018how}.

\subsection{Image Classification}
Applying and adapting machine learning techniques to support solving tasks in the context of preserving cultural heritage is a growing field of research.
Many works address image-based classification of artworks by training an image classifier on the basis of images with known class labels in order to make predictions for images with unknown properties~\cite{fiorucci2020machine}.
First works investigate classical machine learning approaches aiming to predict the characteristics of a depicted painting~\cite{Arora2012}.
In \cite{blessing2010}, one-versus-all Support Vector Machines are trained based on HOG features (histograms of oriented gradients~\cite{dalal2005histograms}) of images showing paintings, with the goal of predicting the artist that created the painting.

Instead of training a classifier to predict variables based on handcrafted image features, Convolutional Neural Networks (CNNs) allow for simultaneously learning features from given input images as well as learning a mapping of these features to class scores based on labeled training images~\cite{LeCun1989, Krizhevsky2012}.
Thus, a trained CNN can be used to predict a class label for an object with unknown properties from an image depicting that object.
CNN-based classifiers are also applied in many works addressing attribute prediction for depicted objects in the context of cultural heritage, where the focus is on making predictions for images showing paintings~\cite{santos2021artificial, castellano2021deep}.
In~\cite{tan2016}, the \textit{artist}, the \textit{genre}, as well as the \textit{style} of a painting, are learned by means of a variant of AlexNet~\cite{Krizhevsky2012}, achieving on average 68.3\% correctly classified images for the three variables in the WikiArt dataset (\url{http://www.wikiart.org/}).
A ResNet18~\cite{he2016deep} was used to predict a painting's \textit{artist} in \cite{sur_blaine_2017} which reported an 82.5\% overall accuracy on the Rijksmuseum dataset~\cite{mensink2014rijksmuseum}.
In both cases, there is one CNN per classification task, and network weights pre-trained on a variant of the ImageNet dataset~\cite{deng2009} are used to improve the classification performance.

Instead of training a separate CNN per task to be learned, the concept of multitask learning aims to exploit interdependencies between related tasks by means of jointly learning them in one network and, thus, to improve the network's performance in solving the individual tasks~\cite{caruana1993}.
Multitask learning for CNNs is addressed in many recent works~\cite{crawshaw2020multi} investigating different strategies for combining the training of several tasks.
In the domain of cultural heritage, the most frequently used strategy applies a feature extraction network producing a high-level image representation that is shared among all tasks and which is processed by additional task-specific layers designed to solve the individual classification tasks~\cite{belhi2018towards, strezoski2017omniart, garcia2020contextnet}.
These works do not only perform multitask learning for predicting characteristics of paintings on the basis of images, but they also make use of pre-trained CNNs for the shared feature extraction network.

In contrast to all works cited so far, which are dedicated to the classification of paintings, we address the CNN-based classification of images of silk fabrics.
Even though there is published work dealing with the CNN-based classification of images of textiles~\cite{xiao2018knitted, puarungroj2019recognizing, iqbal2020woven, meng2021multi}, distinguishing different textile patterns, no work could be found addressing the classification of images of fabrics in the context of cultural heritage except for our previous one.
The image classification network presented in this work can be seen as an expansion of \cite{dorozynski2019multi}, aiming to predict different properties of silk fabrics;
the network takes images of silk fabrics as input, where a high-level image representation produced by a fine-tuned ResNet~\cite{he2016} is shared among all task-specific classification branches that deliver the predictions.
In contrast to \cite{dorozynski2019multi} as well as \cite{belhi2018towards, strezoski2017omniart, garcia2020contextnet}, we adapt the training of the network weights such that hard training examples get a higher impact on the weight updates.
In this way, we want to deal with the problem of class imbalance in the training data, aiming to improve the classification performance for underrepresented classes.
For that purpose, we combine a variant of the focal loss \cite{lin2017focal} with the multitask softmax cross entropy loss used in \cite{dorozynski2019multi}, leading to a training strategy that focuses on hard training examples in a multitask scenario while allowing for missing class labels at training time for some of the tasks to be learned.
Furthermore, we investigate the prediction of four variables instead of three like in \cite{dorozynski2019multi} and evaluate our methodology on a much larger dataset consisting of images from several museum collections instead of only one.

\subsection{Text Classification}
Much of the recent work in natural language processing has focused on fine-tuning large transformer neural networks~\cite{attention} pretrained as language models such as BERT~\cite{bert} and RoBERTA~\cite{roberta}. The most common approach is to add a task-specific head to the pretrained transformer to create the final model architecture. The full model is then trained on the task-specific data. This is the process that is called fine-tuning. On most natural language processing (NLP) tasks, some variation of this approach provides the best results.
Previously, many multilingual and cross-lingual artificial intelligence approaches to text used pretrained and aligned word embeddings, such as the aligned fastText embeddings~\cite{bojanowski2017enriching, joulin2018loss}. But similarly to the overall trend in NLP, recent approaches have also moved towards using fine-tuning of pretrained transformers. Our work follows this trend, we fine-tune the pretrained XLM-R~\cite{conneau-etal-2020-unsupervised}. 
Multitask models are often very desirable from a practical perspective: a single model is easier to deploy and maintain, offers faster inference, and occupies less space in memory when compared to multiple models. Further, multitask learning can often result in measurable improvements \cite{caruana1997multitask}. \cite{liu-etal-2019-multi} showed that multitask training of BERT improved results across several tasks.

The use of NLP, especially text classification, in the cultural heritage domain isn't very widespread. This is a consequence of the fact that the digitization of artifacts usually includes images and some labels or tags, but text descriptions are far less common. The highlight is the work on text descriptions of paintings from the Rijksmuseum Amsterdam described in \cite{ruotsalo2009knowledge}. They used an Information Extraction approach rather than classification. Their pipeline included Named Entity Recognition, Part-of-Speech tagging, and dependency parsing to extract concepts from the text, those concepts were then matched to an ontology and finally classified according to a role. These roles included all the properties we use: Technique, Material, Date, Place, plus others such as "Creator" of the artwork, style, and the subject depicted. Their data, although limited to a total of 250 text descriptions, was manually annotated and each text contained a concept-role pairing. They reported an average F1 of 61.2\% compared to a non-expert human average of 65.1\%. Our work differs from this in several key areas. First, our classification approach is more generalizable, as it does not necessarily require information to be directly present in the text and more resilient to misspellings and non-standard grammar. In fact, even correctly linking information known to represent a certain label (e.g., material) from tables can be challenging in the presence of spelling issues. Second, we work with multilingual data from multiple sources, which presents additional challenges. Finally, our dataset contains many more samples and uses automatic labeling based on information present in catalogs rather than externally annotated.

We are also proposing to improve the handling of class imbalance through the use of the focal loss in text classification. The focal loss was compared to (binary) cross-entropy in the context of multilabel text classification~\cite{huang-etal-2021-balancing}. It was shown to outperform cross-entropy in both the datasets used. The datasets used are significantly different from ours in that they have many more labels and, in particular, many small classes with a very small number of very large classes, forming a "long-tailed" distribution. Thus, the results do not necessarily translate to our data. Further, the text classifier we present uses the multiclass, not the binary, cross-entropy loss, which also differs from their experiments.

\subsection{Tabular Classification}
Gradient Boosted Decision Trees (GBDT)~\cite{friedman2001greedy} has long been the state of the art and the common choice for handling tabular data. Concurrently with our work, Neural Network based alternatives have been proposed which can outperform GBDT in certain situations~\cite{tabnet, kadra2021welltuned}. To the best of our knowledge, there has been no work published on tabular classification in the cultural heritage domain that we can provide an overview of.

\subsection{Multimodal Classification in Cultural Heritage}
A joint image-text neural network architecture for classifying images of paintings by artist and year was presented in \cite{app8101768}. Their text input consisted of a limited set of labels (style, media, and genre) rather than text descriptions. Conceptually, this is similar to CLIP-Art~\cite{Conde_2021_CVPR}, an application of CLIP~\cite{pmlr-v139-radford21a} to the retrieval of artwork images. CLIP learns to associate a small text vocabulary akin to labels with images through joint contrastive pre-training. This was applied to The iMet Collection dataset~\cite{zhang2019imet}, possibly the most similar dataset to our own, it includes images of artworks associated with labels (also called "tags") that describe what is visually depicted in the object (e.g., "Dragons"), its visible properties (dimensions, medium) as well as other culturally relevant properties (e.g., country of origin). Another very relevant dataset in this context is Artpedia~\cite{stefanini2019artpedia}, a dataset of images of paintings associated with textual descriptions tagged as either "visual sentences" that describe the scene depicted in the painting or as "contextual sentences" that describe other aspects of the painting such as its historical context. The tasks for which this dataset was created consist of separating visual from contextual sentences and the retrieval of the correct image for a given text. The differences between the related work and our work are clear. We propose handling images, multilingual text, and tabular data as equal modalities. We also propose to handle data from multiple collections.


\section{Data}

\subsection{Knowledge Graph}\label{sec:data_kg}
The SILKNOW Knowledge Graph lies at the center of all efforts to create a unified representation of the metadata of European silk textiles, particularly from the 15th to the 19th century. The data used in our experiments was collected from 16 sources, most of which are public online museum records, for which we built crawling and harvesting software. In addition to that, we have data from the SILKNOW~\footnote{\url{https://silknow.eu/}} project partners Garin and the University of Palermo (Sicily Cultural Heritage). The dataset used in the experiments was created from a full export of all objects in the knowledge graph, which consists of the metadata of 38,873 unique silk objects before any preprocessing steps. This export includes in total 74,527 unique image files.

In order to model this heterogeneous data from so many sources, we chose and relied strongly on the CIDOC Conceptual Reference Model (CRM). We also developed our own SILKNOW ontology~\footnote{\url{https://ontome.net/namespace/36}} to extend CIDOC-CRM with further classes and properties. This was done for cases where it did not cover some specifics of the silk textile domain and for storing the confidence score of metadata predictions, once we started integrating the results of those predictions into the KG.

In order to develop a converter that could unify all the original data with all these classes and properties into one knowledge base, mappings have been created by domain experts. On a technical level, all museum records had to be harvested and were first converted into a common JSON file format through our crawler software but each array inside this format still had the original field labels from the museums before the final conversion. For example, the majority of museums have a field for describing the production time of a silk object, but in most cases, museums use different names for their field. Moreover, the museums are from all over the world and use different languages for both the field names and their values. This is why we created a mapping for, e.g, a field named "Date" (Metropolitan Museum of Arts) and the class \texttt{E12\_Production} with the property \texttt{P4\_has\_time-span} and another class \texttt{E52\_Time-Span}. Likewise, a mapping rule was written for the field named "date\_text" (API of the Victoria and Albert Museum) and for the (Spanish) field named "Datación" (Red Digital de Colecciones de Museos de España).

Another very central part of our knowledge representation is the SILKNOW Thesaurus, a controlled vocabulary that contains many explicit and multilingual concept definitions for materials, techniques, and motif depictions relevant to these silk textiles. Thanks to this thesaurus, a lot of information and entities from explicit categorical fields of the original museum records could be linked, without any advanced machine learning techniques. The string literal could just be matched with the (multilingual) labels of the thesaurus and then replaced with a unique concept link. This explicit representation of knowledge forms the core of the dataset used to predict missing metadata. This includes cases where a categorical value is either not given at all or “hidden” in longer textual descriptions and not explicitly semantically annotated. 

Once all the modeling, download, conversion, and enrichment steps were taken, the final knowledge graph was uploaded onto a SPARQL endpoint from where all the data across languages and museums can be queried the same way. To make access easier, we also developed a RESTful API, so it is not necessary for web developers to write SPARQL queries, and an aforementioned exploratory search engine on top of this API, called ADASilk. It is aimed at users with only little technical background or little background knowledge about the domain of silk, to make them able to discover a lot of the data in the KG. ADASilk offers advanced search functionality with filters, topic suggestions, and a clean visual interface that shows objects with their images and metadata.

\subsection{Extracting and Normalizing Labels}
\label{sec:data_labels}
The development of the SILKNOW Knowledge Graph is a combined effort of data processing that relies on a data modelling and annotation process created in collaboration with domain experts. This is especially true for the SILKNOW Thesaurus. The group labels used in the experiments in this work are based on the hierarchy and relations of concepts of the silk textile domain described in this controlled vocabulary. As described in Section~\ref{sec:data_kg}, many categorical property values can be easily extracted, linked, and automatically normalized thanks to the SILKNOW Thesaurus. This means that many concepts are accessible even though there were originally different strings, including typos in some cases, synonyms, or written in different languages. An example would be a weaving technique like "Damask", which would be "Damas" in French and "Damasco" in Spanish and Italian: for all these, we replace the string literal with one link to the same concept. In addition to the SILKNOW Thesaurus, we also use linked open data such as GeoNames\footnote{\url{https://www.geonames.org/}} to normalize and link place names. 

\begin{figure*}[htbp]
    \centering
    \includegraphics[width=0.9\textwidth]{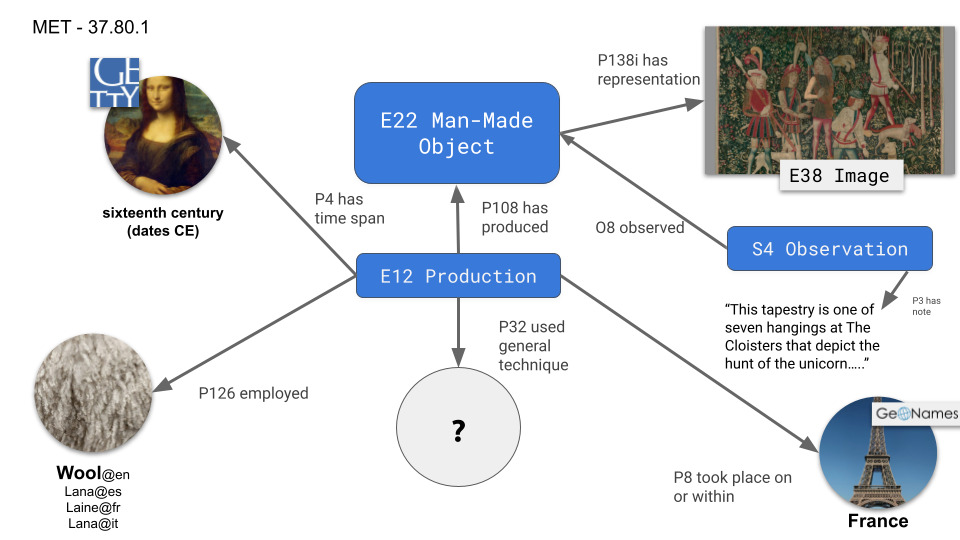}
    \caption{A record from the MET museum with a missing property represented in the knowledge graph using our ontology and controlled vocabularies}
    \label{fig:KG-illustration}
\end{figure*}

Matching strings with our thesaurus and other controlled vocabularies was not without challenges. As will also be discussed in Section~\ref{sec:exp_analysis}, misspellings or unconventional punctuation usage could still cause the matching process to fail. To give an example: If the string value of a record was "silk; gold thread" the latter would not have been linked, due to a bug that did not properly consider a semicolon as a separator. Other such cases existed as well, as the development of the SILKNOW Knowledge Graph is an ongoing process and concurrent with this work. See Figure~\ref{fig:KG-illustration} for an illustration of a museum record in the knowledge graph.

The aforementioned hierarchy defined in the SILKNOW Thesaurus can be used to select specific types or subtypes of properties. To refer back to the previous example, we could select only objects with the weaving technique "Damask", but also only objects made with "Two-coloured damask" which is even more specific. Based on the Thesaurus, we can also make sure that we only choose objects based on equivalent levels of this hierarchy. 

Based on this enrichment and the linking process, we created a pipeline to extract the dataset based on pre-specified criteria. We first developed a comprehensive SPARQL query that outputs all museum objects described in the Knowledge Graph (KG) and includes, if available, the most relevant properties: the identifier of the object in the knowledge graph, the museum where the description comes from, the text description, and URL links to the images that illustrate the object.
The results of this query were exported as a CSV file, which we then post-processed to make sure that we have a format of one row per object. This CSV is used as the basis for all experiments and is distributed as part of our dataset. Table~\ref{tab:dataset_examples} shows examples from this file. In the table, "[NA]" represents a missing value; [URL] replaces URLs; Only small text snippets are included; The ID column was not included.
\tiny
\begin{table*}[htp]
\centering
\caption{Examples of records in our dataset}
\label{tab:dataset_examples}
\begin{tabular}{ccccccc}
\hline\noalign{\smallskip}
Museum & Text & Images & Place & Timespan & Technique & Material \\
\noalign{\smallskip}\hline\noalign{\smallskip}
vam & Furnishing... & [URL1]  	& Great Britain  & XIX 	& [NA]  &  [NA] \\
risd& {[NA]}   		  & [URL1]	    & Japan      	 & [NA] & [NA]  &  Animal fibre \\
met & tapestry... & [URL1-29] & France		 & [NA]  & [NA]  &  Animal fibre  \\
vam & At the end...    &  [URL1-4] & Turkey &      [NA]    & Embroidery & Metal thread \\
\noalign{\smallskip}\hline
\end{tabular}
\end{table*}
\normalsize

\subsection{Label Grouping}\label{sec:label_groups}
In principle, the Knowledge Graph contents can be used to generate training and test samples for the classifiers described in Section~\ref{sec:methods}. 
One would just have to associate the images and/or the text given for a record with the annotations in the categorical variables of interest. The available annotations can be easily converted into class labels.
However, a statistical analysis of these annotations revealed that most of them occur very rarely in the data, while for all categorical variables, there were one or a few classes that were dominant in the sense that many records belonged to them. 
Supervised classifiers have problems with imbalanced training data sets, and it would seem very difficult for a classifier to successfully differentiate classes for which it has seen only a very small number of training samples if on the other hand, there are thousands of samples for some other classes. 
To still be able to extract meaningful information from the available modalities using supervised methods while at the same time having the chance of achieving a reasonably good classification performance, a simplified class structure was defined. 
Domain experts analyzed the class distributions and aggregated classes corresponding to different categories into compound classes. Care was taken for the aggregated classes to be consistent with the Thesaurus, and aggregated only if they were considered to be related according to the domain experts. 
At the same time, the aggregation was guided by the frequency of occurrence of class labels so that the compound classes would occur frequently enough to be used for training the supervised classifiers described in Section~\ref{sec:methods}. 

The resultant simplified class structure was integrated into the Knowledge Graph in the form of what we called {\em group} fields, which were made available for all semantic properties of interest, principally, the ones corresponding to the different tasks in this work. Such grouping was applied to the following properties: Material, Technique, production place (with a country granularity), production time (with the century granularity), and the object type or object domain group, to be able to filter out non-textiles that use silk. Grouping was not an easy task, domain experts had to deal with more than 200 concepts that had to be grouped according to the aforementioned categories. Techniques were the most complex to group. To do so, domain experts grouped the concepts according to two fundamental criteria: 1) whether they belonged to the same hierarchy, for example, velvet and its types. In fact, there are many types of velvet, classified depending on the nature of the pile such as broderie velvet, ciselè velvet, cut velvet, pile-on-pile velvet, uncut velvet, etc. 2) If they were somehow related to a certain technique, for example, the effects obtained of applying differently warp and weft, that is, whenever a yarn is introduced into a fabric to produce an effect or pattern.  On the other hand, materials were not complex as they were made in large groups according to their origin, that means according to the product obtained from the processing of one or more raw materials, in the course of which their structure has been chemically modified, e.g., animal fibres are distinguished from vegetable fibres.
Using a conversion table for aggregation prepared by the domain experts, the contents of the {\em group} fields could be derived automatically from the original semantic annotations. 
Having thus expanded the Knowledge Graph, training, and test samples could be easily generated from it by appropriate SPARQL queries that would export the contents of the {\em group} fields associated with each record. 

\subsection{Dataset Preparation and Properties}\label{sec:data_split}
The goal of the dataset preparation is the conversion of the knowledge graph data with normalized and grouped labels described, respectively, in Sections \ref{sec:data_kg}, \ref{sec:data_labels}, and~\ref{sec:label_groups}, into a dataset for the experiments in Section \ref{sec:exp_analysis} using the classification methods described in Section \ref{sec:methods}. 

The first step was to select the records in the knowledge graph that were relevant to the domain. The second step was to select only records that contained a value for one of the variables to be predicted, i.e., labeled samples. Uncommon labels, with a total frequency below 150, were discarded. The final step was to randomly split the records into disjoint sets: 

\begin{itemize}
    \item a training set consisting of 60\% of the data for supervised learning;
    \item a validation (or development) set, consisting of 20\% of the data for hyperparameter tuning and multimodal supervised learning;
    \item a test set, consisting of 20\% of the data, for the final evaluation of the proposed methods.
\end{itemize}

Given that the objective is to train and evaluate a multimodal multitask approach on records, that regularities exist within each collection (i.e., museum) that comprises the data, that the text modality is also multilingual, and that both modalities and task-specific labels may be missing from a record, we believe the most reasonable way to split the dataset is a random split of records. The distribution of the data, in terms of records per variable, set, and class label, sorted by count, can be seen in Table~\ref{tab:class_stats}. The table includes separate columns with the count of all records, columns for records with images, and columns for records with text. The first can be used by the tabular classifier, the second by the image classifier, and the third by the text classifier. Because records can have multiple images, while they may or may not have text, the exact class proportions can change depending on the modality. Overall, class imbalance exists for all variables.

\begin{table*}[htp]
\centering
\caption{Class structure and class distribution of the records}
\label{tab:class_stats}
\begin{tabular}{lc>{\columncolor[gray]{0.8}}cc|c>{\columncolor[gray]{0.8}}cc|c>{\columncolor[gray]{0.8}}cc}
\hline\noalign{\smallskip}
& \multicolumn{3}{c}{TRAIN} & \multicolumn{3}{c}{VALIDATION}  & \multicolumn{3}{c}{TEST} \\
&       & with  & with        &      & with   & with      &       & with   & with \\
&  total& images& text        &total & images & text      & total & images & text \\
\hline\noalign{\smallskip}
\multicolumn{10}{l}{\textit{place}} \\ 
FR      & 3,156 & 3,115 & 862       & 1,037 & 1,030 & 312       & 1,072 & 1,062 & 290 \\
IT      & 1,853 & 1,827 & 314       & 687   & 682 & 113       & 665   & 654 & 117 \\
GB      & 1,721 & 1,697 & 1,327     & 562   & 548 & 428       & 554   & 543 & 437 \\
ES      & 1,605 & 1,550 & 973       & 521   & 503   & 313       & 504   & 493 & 293 \\
IN      & 735   & 723 & 612       & 231   & 228   & 197       & 224   & 221 & 195 \\
CN      & 426   & 362   & 232       & 127   & 107   & 79        & 146   & 120 & 83 \\
IR      & 409   & 377   & 269       & 142   & 131   & 94        & 120   & 111 & 82 \\
JP      & 325   & 276   & 120       & 92    & 82   & 44        & 116   & 95 & 43 \\
TR      & 205   & 203   & 145       & 57    & 57    & 39        & 69    & 68 & 46 \\
\noalign{\smallskip}\hline\noalign{\smallskip}
\multicolumn{10}{l}{\textit{timespan}} \\
XIX     & 3,492 & 3,291 & 2,268     & 1,180 & 1,115 & 735       & 1,177 & 1,112 & 761 \\
XVIII   & 2,576 & 2,531 & 1,058     & 901   & 890 & 371       & 920 & 901 & 325 \\
XX      & 1,520 & 1,428 & 749       & 483   & 452   & 220       & 480 & 452 & 224 \\
XVII    & 689   & 679 & 392       & 231   & 228   & 113       & 214 & 209 & 118 \\
XVI     & 542   & 536 & 236       & 180   & 178   & 75        & 158 & 157 & 71 \\
\noalign{\smallskip}\hline\noalign{\smallskip}
\multicolumn{10}{l}{\textit{technique}} \\
embroid. & 1,814 & 1,788 & 662     & 657 & 652 & 198       & 652 & 636    & 203 \\
velvet    & 1,273 & 1,265 & 159     & 454 & 451   & 54        & 466 & 464    & 57 \\
damask    & 1,004 & 977 & 422     & 333 & 323   & 144       & 348 & 343    & 135 \\
other     & 722   & 710 & 351     & 219 & 215   & 97        & 209 & 207    & 92 \\
\noalign{\smallskip}\hline\noalign{\smallskip}
\multicolumn{10}{l}{{\textit{material}}} \\

animal    & 10,387& 9,938 & 2,544  & 3,445 & 3445 & 793     & 3,550 & 3,550 & 797 \\
vegetal   & 1,255 & 1,226  & 674    & 396   & 387   & 207     & 400   & 391   & 187 \\
metal t. & 1,223 & 1,208  & 382    & 422   & 421   & 135     & 401   & 396   & 116 \\
\noalign{\smallskip}\hline
\end{tabular}
\end{table*}

An additional overview of the modalities with their overlap can be found in Table~\ref{tab:modality_stats}. We can see how 27.120 or 96,60\% of the 28,077 records about annotated fabric objects contain at least one image, but only 11,034 or 39,29\% of them contain a text description. The overlap consists of 10,664 or 37,98\%. The proportion between training validation and test sets in each case corresponds roughly to the aforementioned 60-20-20 split. 
Finally, the distribution of samples over the museums can be found in Table~\ref{tab:museum_stats}. 

\begin{table}[htbp]
\centering
\caption{Distribution of records per museum}
\label{tab:museum_stats}
\begin{tabulary}{\linewidth}{Llr}
\hline\noalign{\smallskip}
Museum name &  ID & records \\
\noalign{\smallskip}\hline\noalign{\smallskip}
Metropolitan Museum of Arts &met & 6,524 \\
CDMT Terrassa&imatex & 6,119  \\
Victoria and Albert Museum &vam & 5,527 \\
Rhode Island School of Design&risd & 3,226 \\
Boston Museum of Fine Arts &mfa & 2,610 \\
Garín 1820 &garin & 1,558  \\
Collection du Mobilier National&mobilier & 1,293 \\
Red Digital de Colecciones de Museos de España &cer & 781  \\
 Joconde Database of French  Museum Collections &joconde & 375  \\
Smithsonian Museum &smithsonian & 38  \\
Versailles &versailles & 18  \\
Art Institute of Chicago &artic & 8  \\
\noalign{\smallskip}\hline
\end{tabulary}
\end{table}

\begin{table}[htbp]
\centering
\caption{Modality statistics and overlap}
\label{tab:modality_stats}
\begin{tabular}{lrrrrr}
\hline\noalign{\smallskip}
data       & with   & with  & both  & neither  \\
split      &  image & text     &   &  \\
\noalign{\smallskip}\hline\noalign{\smallskip}
train      & 16,260   &  6,717   &  6,495   & 358 \\
val.        &  5,419   &  2,184   &  2,101   & 100\\
test        &  5,441   &  2,133   &  2,068   & 129 \\
\hline
total   & 27,120   & 11,034   & 10,664   & 587 \\
        & 96.6\% & 39.3\%  & 38.0\% & 2.1\% \\
\noalign{\smallskip}\hline
\end{tabular}
\end{table}

Each record in our dataset corresponds to an object in a museum. Each record can have multiple images associated with it, but at most, one "text". Text data in our dataset consists of descriptions of fabrics or objects made mostly of fabrics. These descriptions range in length from short sentences to multi-sentence paragraphs to multi-paragraph texts with thousands of words. Some descriptions focus primarily on a single aspect, such as a scene depicted or the history of the object, while others focus on various properties of the object. In order to eliminate some errors present in the data, such as placeholder text and web errors, we removed any text descriptions smaller than 60 characters. The resulting distribution of lengths, in terms of characters and space-delimited tokens (roughly, words), is summarized in Table~\ref{tab:text_lens}. These descriptions are in 4 different languages: English, Spanish, French, and Catalan. The counts for each are shown in Table~\ref{tab:text_lang}

\begin{table}[htbp]
\centering
\caption{Text length in characters and tokens}
\label{tab:text_lens}
\begin{tabular}{lcccccc}
\hline\noalign{\smallskip}
& Min & Q1 & Med. & Mean & Q3 & Max \\
\noalign{\smallskip}\hline\noalign{\smallskip}
Char & 60 & 173 & 343 & 693 & 856 & 16,333 \\
\noalign{\smallskip}\hline\noalign{\smallskip}
Token & 7 & 28 & 56 & 115 & 142 & 2,826 \\
\noalign{\smallskip}\hline
\end{tabular}
\end{table}

\begin{table}[htbp]
\centering
\caption{Language distribution of text}
\label{tab:text_lang}
\begin{tabular}{lcccc}
\hline\noalign{\smallskip}
& English & Spanish & French & Catalan \\
\noalign{\smallskip}\hline\noalign{\smallskip}
Records & 7271 & 1975 & 1126 & 680 \\
\noalign{\smallskip}\hline
\end{tabular}
\end{table}


\section{Methods}
\label{sec:methods}


\subsection{Image Classification}
\label{sec:image_classifier}


The goal of the image classification is to predict one class label per classification task, i.e., the prediction of a class label for each of the target variables \textit{technique}, \textit{timespan}, \textit{material}, and \textit{place}, for an image that illustrates an object. 
For that purpose, an image classifier is trained using all images of all records contributing to the dataset described in Section~\ref{sec:data_split}. 
We propose to use a convolutional neural networks (CNN) for that purpose, motivated by the success of CNN in image classification. 
As there are many records with annotations for more than one of these variables, we propose to train the classifier to predict all classes simultaneously in a multitask framework, exploiting the inherent relations between the variables to learn a joint representation that is used by task-specific classification heads. 
A detailed description of the chosen network architectures can be found in Section~\ref{sec:methods_image_architecture}, whereas the strategies used for training are presented in Section~\ref{sec:methods_image_training}.

\subsubsection{Network Architecture}
\label{sec:methods_image_architecture}

\begin{figure*}[ht]
  \centering
  \includegraphics[width=0.6\paperwidth]{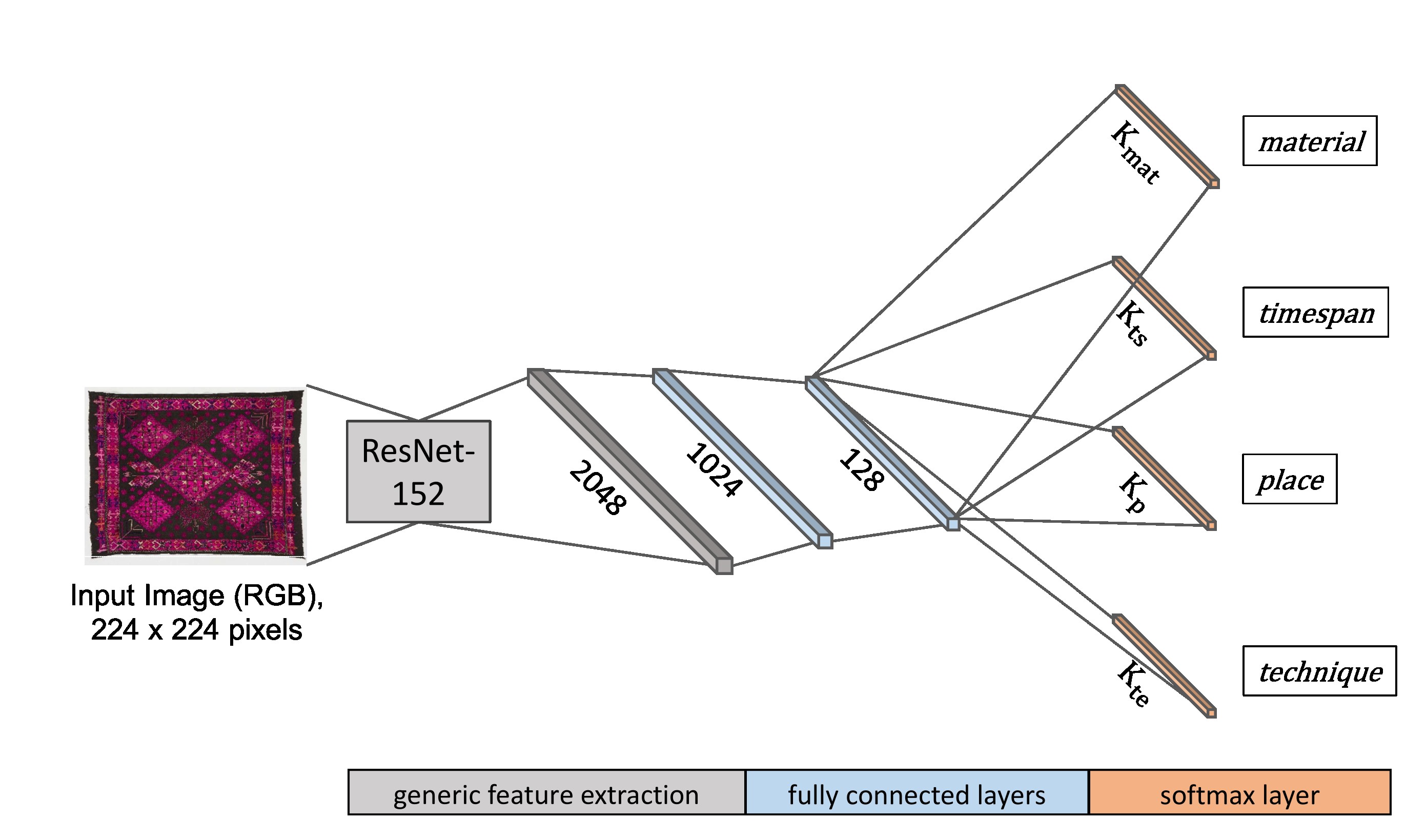}
\caption{Network architecture of the CNN for multitask image classification}
\label{fig:mtl_cnn_image}
\end{figure*}

Figure~\ref{fig:mtl_cnn_image} shows the structure of the CNN for multitask learning for the prediction of the four target variables. 
Its input consists of an RGB image scaled to a size of 224 x 224 pixels. 
This image is presented to the ResNet 152 network~\cite{he2016} pre-trained on ImageNet~\cite{deng2009}, which serves as a generic feature extractor for the image~\cite{sharif2014} and produces a feature vector of 2048 dimensions. 
We apply dropout with a probability of 10\% after this layer~\cite{Srivastava2014}. 
This is followed by $L_{fc}  = 2$ fully connected layers, the first one having 1024 and the second one having 128 nodes, which are shared by all tasks.
Rectified linear units~\cite{nair2010rectified} are used as nonlinearities in both of these joint layers. 
They produce a joint representation of the image of $N_{r} = 128$ dimensions. 
This representation is processed by four task-specific classification branches, each consisting of one additional softmax layer only, which delivers the class scores $y_{km}\left (\mathbf{x}, \mathbf{w}\right)$ for the input image $\mathbf{x}$ to belong to class $k$ for variable $m$. 
The number of nodes of the softmax layer corresponds to the number of classes to be differentiated for a specific task. In Figure~\ref{fig:mtl_cnn_image}, these are represented by \textit{$K_{mat}$}, \textit{$K_{ts}$}, \textit{$K_{p}$}, and \textit{$K_{te}$} corresponding to the tasks \textit{material}, \textit{timespan}, \textit{place}, and \textit{technique}, respectively. 

The CNN predicts one class label per task for every image. In the case of multiple images per record, one such class label is predicted for each one of the images, and the prediction with the highest softmax score is chosen to be the prediction for the record. 

\subsubsection{Training}
\label{sec:methods_image_training}
During training, the parameters $\mathbf{w}$ of the CNN described in Section~\ref{sec:methods_image_architecture} are learned by minimizing a loss function $E(\mathbf{w})$. 
The parameters of our network consist of the parameters $\mathbf{w}_{R}$ of ResNet-152, which are initialized from the published pre-trained model~\cite{he2016}, and the parameters $\mathbf{w}_{FC}$ of the fully connected and softmax layers, which are initialized randomly by a variant of the Xavier initialization~\cite{he2016}.
In the training procedure, we determine the parameters $\mathbf{w}_{Rt}$ of the last $NL_{RT}$ layers of ResNet-152 considering exclusively entire residual blocks and the parameters $\mathbf{w}_{FC}$ of the fully connected layers, whereas the parameters $\mathbf{w}_{Rf}$ of the first $152-NL_{RT}$ ResNet-152 layers are frozen~\cite{Yosinski2014}. 
Thus, the parameter vector consists of three subsets: $\mathbf{w} = \left(\mathbf{w}_{Rf}^T, \mathbf{w}_{Rt}^T, \mathbf{w}_{FC}^T\right)^T$. 
$NL_{RT}$ is a hyperparameter to be tuned. 

Two loss functions can be used for training the network. 
The first one is an extension of the standard softmax cross-entropy loss~\cite{BISHOP2006} with weight decay~\cite{dorozynski2019multi}:

\begin{align}
	&E_{SCE}\left(\mathbf{w}\right) = \nonumber \\
	&- \sum_{n=1}^{N}\left(\sum_{m \in M_n} \sum_{k=1}^{K_m}  t_{nmk}\cdot ln\left(y_{km}\left(\mathbf{x}_n, \mathbf{w}\right)\right)\right) \nonumber \\
	&+ \omega_R\cdot R\left( \mathbf{w}_{Rt}, \mathbf{w}_{FC} \right) 
\label{eq:loss_incomp_MTL}
\end{align}

In eq.~\ref{eq:loss_incomp_MTL}, $y_{km}\left(\mathbf{x}_n, \mathbf{w}\right)$ is the softmax score for the $n^{th}$ training image $\mathbf{x}_n$ to belong to class $k$ for variable $m$. 
The indicator variable $t_{nmk}$ is one if the class label of sample $n$ for variable $m$ is $k$ and zero otherwise. 
The sum is taken over all $N$ training samples and $K_m$ classes for task $m$. 
$M_n$ is the set of tasks for which the true class label is known for the training sample $n$, so that the loss in eq.~\ref{eq:loss_incomp_MTL} considers exclusively samples $x_n$ with $t_{nmk}=1$ for learning task $m$. 
In this way, the fact that the annotations for most samples are incomplete, i.e. that annotations are only available for a subset of the variables to be predicted, can be considered.
If multiple annotations are available, the corresponding classification losses will be backpropagated to the joint layers from multiple classification branches, thus supporting the learning of a joint representation for all variables. 
The outputs for variables for which the true class label is unknown will not contribute to the loss and to the parameter update. 
Finally, the term $R\left( \mathbf{w}_{Rt}, \mathbf{w}_{FC} \right)$ corresponds to regularization by weight decay, which is only applied to the parameters to be updated in training; $\omega_R$ is a hyperparameter defining the influence of this term on the result. 

One problem of the data described in Section~\ref{sec:data_split} is its imbalanced class distribution. 
In this case, minimizing the cross-entropy loss in eq.~\ref{eq:loss_incomp_MTL} will favor the dominant classes, resulting in a poor performance for the underrepresented ones. 
In order to mitigate these problems, a multi-class extension of the focal loss~\cite{lin2017focal,liu2018age} with regularization is utilized for training: 

\begin{eqnarray}\label{eq:loss_incomp_MTL_focal}
	E_F\left(\mathbf{w}\right) = \nonumber \\ 
	- \sum_{m \in M_n}\left(\sum_{n=1}^{N} \sum_{k=1}^{K_m}  \left(1- y_{km}\left(x_n, \mathbf{w}\right)\right)^{\gamma} \right. \nonumber \\
	\Biggl. \cdot t_{nmk}\cdot ln\left(y_{km}\left(x_n, \mathbf{w}\right)\right) \Biggr) \nonumber \\
	+ \omega_R\cdot R\left( \mathbf{w}_{Rt}, \mathbf{w}_{FC} \right)
\end{eqnarray}

The only difference between the loss functions in eqs.~\ref{eq:loss_incomp_MTL} and~\ref{eq:loss_incomp_MTL_focal} is the penalty term $\left(1- y_{km}\left(x_n, \mathbf{w}\right)\right)^{\gamma}$, where $\gamma$ is a hyperparameter modulating the influence of this term on the result. 
This penalty term forces the loss to put more emphasis on samples that are difficult to classify (having a small score $y_{km}$ for the correct class). 
Assuming the samples of underrepresented classes to be hard to classify by the CNN, this loss is expected to improve the results for these classes. 

Starting from initial values derived in the way described earlier, stochastic minibatch gradient descent based on the Adam optimizer~\cite{kingma2014adam} is applied to determine the CNN parameters, using the default parameters ($\beta_1=0.9$, $\beta_2=0.999$, $\varepsilon=10^{-8}$) and a minibatch size of 300. 
The base learning rate $\eta$ is another hyperparameter to be tuned. 
We use early stopping 
and use the model parameters leading to the lowest loss on the validation set.

\subsection{Text Classification}
\label{sec:text_classifier}
Our problem is defined as value prediction for certain properties of an object, a silk fabric, given its text description, which can be written in any one of the four languages listed in Table \ref{tab:text_lang}. We have 4 tasks, each denominated according to the property of the underlying fabric object we want to predict: the \textit{technique} and \textit{material} used to create it, the \textit{timespan} or time period when it was created, and the \textit{place} where it was created. While some descriptions directly contain some of this information, as seen in Table~\ref{tab:text_examples}, this is sufficiently uncommon to prevent a purely extractive approach from yielding good results. For example, of the 4 texts we showed, only one gives a direct indication as to where it was produced (\textit{"British"}). We instead rely on regularities present in the text descriptions to make informed guesses. More technically, we frame our problem as a multiclass, multitask, multilingual text classification problem. That is, given a text description of a fabric, written in any language, we want to assign exactly one label out of a set of mutually exclusive class labels for each of the properties we wish to predict, i.e., the tasks. 

\begin{figure}[!ht]
  \centering
  \includegraphics[width=0.28\paperwidth]{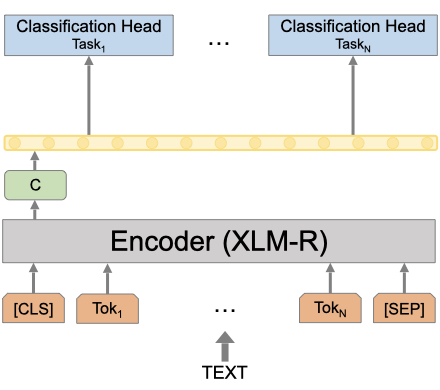}
  \caption{Multitask architecture of the text classifier consisting of a shared XLM-R based encoder followed by task-specific classification heads}
  \label{fig:text_arch}
\end{figure}

\begin{figure}[!ht]
  \centering
  \includegraphics[width=0.15\paperwidth]{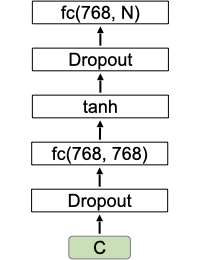}
  \caption{Task-specific classification head of the text classifier consisting of a fully connected (FC) layer followed by a tanh activation followed by the output projection FC layer}
  \label{fig:text_head}
\end{figure}

The text classifier uses a hard parameter sharing based multitask architecture~\cite{ruder2017overview}, shown in Figure~\ref{fig:text_arch}. It consists of a shared encoder followed by task-specific classification heads. The encoder is the multilingual large pretrained transformer, XML-R~\cite{conneau-etal-2020-unsupervised}. Following the method outlined in~\cite{bert}, a special classification token, "CLS", is prepended to all inputs. The final hidden state corresponding to this token, "C", is used as the aggregate sequence representation. It is the only transformer output forwarded to the classification heads. All classification heads are identical except for the output dimension of the last layer, the output projection layer, which equals the number of classes of the task. A diagram of a classification head is shown in Figure~\ref{fig:text_head}. It consists of a fully connected (FC) layer followed by a tanh activation, followed by the output projection FC layer. Dropout is applied before both FC layers with a 10\% probability. A softmax function can convert the output logits of the last layer to normalized probabilities. 

To finetune our transformer-based classifier, at each step, a task is randomly selected using proportional sampling. A batch of examples for this task is then created and fed to the classifier. The cross-entropy loss is then calculated and weights are adjusted through backpropagation. Adam~\cite{kingma2014adam} is used as the optimizer with weight decay~\cite{LoshchilovH19}. For dealing with class imbalance, we will show results with the multiclass extension of the focal loss~\cite{lin2017focal,liu2018age} as an alternative to the cross entropy loss.

\subsection{Tabular Classification}
\label{sec:tab_classifier}
When considering Knowledge Graph records of objects, we can represent them as structured data. That is a table where each row represents an object and each column a property. We use four separate task-specific classifiers to perform tabular classification. These all use the same learning algorithm, Gradient Boosted Decision Trees (GBDT)~\cite{friedman2001greedy}, implemented in XGBoost~\cite{xgboost}. The input to the tabular classifier consists of the categorical values of non-target variables plus the identifier for the museum, as shown in Table~\ref{tab:tab_input}. We replace missing values for a feature with a predefined value, represented by the symbol "[NA]" ("Not Available") in the table. The output of the classifier, for each example, consists of an N-dimensional logit vector. It is used with the softmax function to predict a target class out of N possible classes. This classifier is trained by gradient descent to minimize the softmax loss.

\begin{table*}[hbt]
\caption{Tabular Classification input with one example row per task}
\label{tab:tab_input}
\centering
\begin{tabular}{lcccccc}
\hline\noalign{\smallskip}
Target      & Target            & \multicolumn{5}{c}{Feature}  \\
Variable    & Value             &  museum   & place     & timpespan     & technique         & material\\
\noalign{\smallskip}\hline\noalign{\smallskip}
place       & FR                & risd      & -         & [NA]          & [NA]              & animal fibre\\
timespan    & XVIII             & met       &   [NA]    & -             & embroidery        & animal fibre \\
technique   & other             & garin     &   ES      & XX            & -                 & vegetal fibre \\
material    & vegetable fibre   & vam       &   GB      & XIX           & embroidery        & - \\
\noalign{\smallskip}\hline
\end{tabular}
\end{table*}

\subsubsection{Hyperparameters}
While a detailed explanation of each hyperparameter that controls the resulting model and learning of GBTs is beyond the scope of this work, we believe some contextualization is required. This is due to the relatively larger number of hyperparameters tuned for GBTs in Section~\ref{sec:exp:tab} compared to the Neural Network based methods used for the other modalities, and for the convenience of the reader. 

The hyperparameters \textit{max\_depth} (maximum depth of a tree), \textit{min\_child\_weight} (minimum weight for tree partitioning), and \textit{gamma} (minimum loss reduction for tree partition) all directly control model complexity, which in turn can have significant consequences in terms of fitting.
The hyperparameters \textit{subsample} (the percentage of data sampled per iteration) and \textit{colsample\_bytree} (the ratio of features sampled per iteration) can reduce overfitting by adding random noise to the iterative tree-building process.
Finally, the \textit{learning rate} and \textit{number of rounds} control, respectively, the amount of learning per round and the total amount of learning (i.e., the total number of trees).
"Sample weight" is an optional vector that assigns to each training sample a weight. We either use no sample weighting or assign to each sample a weight according to its class, using the heuristic discussed in Section~\ref{sec:balance}, effectively implementing weight rescaling.

\subsection{Multimodal Classification}
\label{sec:multi_classifier}
\begin{figure}[!ht]
  \centering
  \includegraphics[width=0.3\paperwidth]{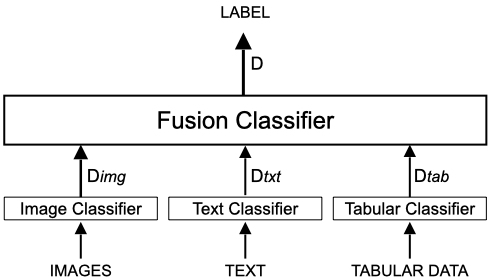}
  \caption{Architecture of the multimodal classifier}
  \label{fig:multimodal_arch}
\end{figure}

Our approach to multimodal classification, shown in Figure~\ref{fig:multimodal_arch}, follows a decision level late fusion approach, in which the decision (prediction), $D_{c}$, from each of the 3 modalities serves as the input to a classifier that makes the final decision of which task-specific label to assign to the record.  We choose the GBDT algorithm for the multimodal classifier.  The input is just one column for each of the three modalities, each column containing the class labels predicted by the corresponding classifier for the task. If a modality is missing, the values in the corresponding column are set to the missing value indicator [NA], just in the way missing class labels are considered by the tabular classifier. Thus, the multimodal classifier can cope with incomplete records (i.e., records with missing modalities) by design. 
We created a separate multimodal classifier for each task, i.e., no multitask learning is applied in multimodal classification.

There are several advantages to late fusion over early or intermediate-level fusion in our case. Firstly, each record may have multiple images but a single text description. Effectively, the input dimensionality is different. With late fusion, we allow the image classifier to deal with it independently, e.g., by classifying multiple images for the same object and picking the decision with the highest confidence. Secondly, the decisions, represented by a one-hot class vector, have a smaller dimensionality than intermediate representations and thus are more appropriate for scenarios with few samples, which is a common problem in the cultural heritage domain. 

\subsection{Methods for dealing with class imbalance}
\label{sec:balance}
As already discussed, our dataset has an imbalanced class distribution. We have already proposed the use of the focal loss for the image and text classifiers. To serve as a comparison, we implemented two other common strategies for dealing with class imbalance. These are Weight Rescaling (WR) and Random Uniform Sampling (RUS). We will evaluate them on the validation set. In the case of the image and text classifiers, they will be subject to separate hyperparameter tuning and shown in separate tables to provide a comparison to the proposed use of the focal loss. In the case of the tabular and multimodal classifiers, weight rescaling will be part of hyperparameter tuning, as we're not proposing an alternative. 

Weight rescaling works by assigning a weight to each class in the cross-entropy loss function. These weights are calculated in such a way as to counter the imbalance present in the data as determined by the training data distribution. We used the common "balanced" heuristic in Scikit-learn~\cite{scikit-learn} attributed to \cite{king2001logistic}. This assigns to each class a weight, $w_c$, given by 
\begin{equation}\label{eq:balanced_class_weight}
	\mathbf{w}_{c} = \frac{N}{C * N_c}
\end{equation}

where $N$ is the total number of samples, $C$ is the number of classes, and $N_c$ represents the number of samples with that class. These weights are calculated independently for each task. These weights are then used to scale the (cross-entropy) loss for each sample during training, according to its true class. We evaluate the performance of WR in all classifiers.

Random Uniform Sampling selects training examples uniformly from each class so that the algorithm sees approximately the same number of examples for each class. To do so, it must repeat the examples of less common classes more often than those of more common classes. We combine random uniform class selection with random uniform task selection. This strategy is only evaluated on the image and text classifiers. The reason is that the GBDT algorithm used by the tabular classifier and multimodal classifier has its own internal sampling logic.

\subsection{Evaluation Metrics}
\label{sec:methods:metrics}
Throughout Section~\ref{sec:exp}, we will report experimental results using two metrics: F1 and accuracy. When choosing between different models or hyperparameters, we use the F1 metric, evaluated over the validation set. When showing experiment results for the classifiers, we also show accuracy to provide more information.

The overall accuracy $OA$ describes the percentage of correctly classified examples, denoted as true positives $TP$, among all classified examples:
\begin{equation}
accuracy = \frac{\text{Correct classifications}}{\text{All classifications}}
\end{equation}
As the $OA$ is biased towards classes with more examples in an imbalanced class distribution, the classification performance of underrepresented classes is not reflected by the $OA$.
In contrast, the class-specific F1 scores, being the harmonic means of precision (i.e., the percentage of the examples assigned to a certain class that actually corresponds to that class in the ground truth, eq.~\ref{eq:prec}) and recall (i.e., the percentage of the examples of a class according to the ground truth which is also assigned to that class by the model, eq.~\ref{eq:rec}) reflect the classifier's ability to predict a certain class:

\begin{eqnarray}
F_1 = 2\frac{precision \cdot recall}{precision + recall} \label{eq:f1}\\
precision = \frac{TP}{TP + FP} \label{eq:prec}\\
recall = \frac{TP}{TP + FN} \label{eq:rec}
\end{eqnarray}

Where false positives ($FP$), correspond to examples of another class incorrectly assigned to the class and false negatives ($FN$) correspond to examples of the class not assigned to it. We report the average F1 scores (also referred to as macro-averaged F1 score, $maF_{1}$) per variable, i.e., the average values of all class-specific $F_{1_i}$ scores of the classes $C$ for that variable: 

\begin{equation}
maF_{1} = \frac{1}{C} \sum_{i=1}^{C} F_{1_{i}}
\end{equation}

This metric gives equal weight to all classes, regardless of the number of samples in each. It is suitable for use in imbalanced settings in which all classes are considered equally important. This is the setting considered in this work, although it is possible to conceive of other work in this domain where different classes have different domain-specific importance weights. In such a case, a weighted macro-averaged F1 score might be considered instead.

We also average the F1 scores of all variables (tasks) to provide a single-number summary for the experiment. This again is an unweighted average that does not take into account the number of samples in each variable (task).

Finally, in Section~\ref{sec:exp_analysis} we additionally show confusion matrices for each modality. A Confusion matrix $M$ shows in $M_{i, j}$ how many examples of class $i$, according to the ground truth, are predicted by the classifier to be of class $j$, thus all values outside the diagonal i.e., where $j \neq i$, are counts of misclassified examples.


\section{Experiments and Results}\label{sec:exp}

\subsection{Image Classification}\label{sec:exp:image}
For all experiments in the frame of image classification, we use the split of the dataset described in Section~\ref{sec:data_split} in order to train the CNN for image classification presented in Section~\ref{sec:methods_image_architecture} by means of the training strategy described in Section~\ref{sec:methods_image_training}.
We use all images that are assigned to a record for training and classification, assigning the class labels of the corresponding records to all images associated with it. As pointed out in Section~\ref{sec:methods_image_architecture}, for records associated with multiple images, all images are classified by the CNN at test time, and the image-based prediction having the highest class score is chosen to be the final result.

\paragraph{Experimental setup.}
The workflow of our experiments is as follows:
The training dataset is used to update the weights $\left(\mathbf{w}_{Rt}^T, \mathbf{w}_{FC}^T\right)^T$ of the CNN with early stopping. The model parametrization and hyperparameters leading to the lowest loss are calculated on the validation set.

In this context, we compared different training strategies dealing with class imbalance discussed in Section~\ref{sec:balance}: the focal loss, weight rescaling, and random uniform sampling.
Furthermore, we tuned the hyperparameters listed in Table~\ref{tab:hyper_parameters_image} and described in Section~\ref{sec:image_classifier}, choosing the values achieving the highest average F1 scores on the validation set. 

Finally, all test set records for which at least one image is available are used for an independent evaluation, with the values of the hyperparameters tuned on the validation set.

\begin{table}[h!]
\centering
\caption{Hyperparameters tuned (image classification)
}
\label{tab:hyper_parameters_image}
\begin{tabulary}{\linewidth}{Lcc}
\hline\noalign{\smallskip}
Hyperparameter & Range & Best \\
\hline\noalign{\smallskip}
Learning Rate $\eta$                    & [1e-5, 1e-3]  & 1e-4  \\
Weight Decay $\omega_R$                 & [0.0, 1e-5]   & 1e-3  \\
Degree of fine-tuning $NL_{RT}$                              & [0, 36]       & 30\\

\noalign{\smallskip}\hline
\end{tabulary}
\end{table}

We will report the overall accuracies as well as the average F1 scores of the best CNN variant in terms of the average F1 score obtained on the test and validation sets for two variants: the first CNN variant is trained by minimizing the softmax cross-entropy loss (equation \ref{eq:loss_incomp_MTL}), whereas the second variant is trained by minimizing the focal loss (equation \ref{eq:loss_incomp_MTL_focal}).

\paragraph{Results.}
Table~\ref{tab:MTL_dev_train_strategy} shows the results obtained by the different methods for handling class imbalance, as well as the standard softmax loss (i.e., the de facto baseline), on the validation set. Applying the focal loss (eq.~\ref{eq:loss_incomp_MTL}) provides the best results compared to all other methods. The next best results were obtained by uniform sampling, while weight rescaling performed worse than the baseline.

\begin{table}[h!]
 \centering
\caption{Comparison of strategies for dealing with class imbalance showing average F1 scores (F1) and average overall accuracies (OA) on the validation set (image classification) 
}
\label{tab:MTL_dev_train_strategy}
\begin{tabular}{lcc}
\hline\noalign{\smallskip}
 & F1  & OA  \\
\noalign{\smallskip}\hline\noalign{\smallskip}
Focal loss (F)              & 58.8  & 71.5    \\
\noalign{\smallskip}\hline\noalign{\smallskip}
Random Uniform Sampling (U) & 56.4  & 67.0 \\
\noalign{\smallskip}\hline\noalign{\smallskip}
Softmax loss (S)            & 55.4  & 70.2    \\
\noalign{\smallskip}\hline\noalign{\smallskip}
Weight Rescaling (W)        & 53.8  & 59.0    \\
\noalign{\smallskip}\hline\noalign{\smallskip}
$\Delta$ (F - U)            & 2.4  & 4.5   \\
$\Delta$ (F - S)            & 3.4  & 1.3   \\
$\Delta$ (F - W)            & 5.0  & 12.5  \\

\noalign{\smallskip}\hline
\end{tabular}
\end{table}

Table~\ref{tab:MTL_dev_test_eval_object} shows results, broken down by task, in both the validation and test sets, comparing the focal loss variant with the standard softmax variant again serving as the baseline. These quality metrics are determined on the basis of the prediction results for records (i.e., not on the raw results for individual images in case of records having multiple images). In this section, some general observations and the conclusions drawn from them will be briefly described, and a more detailed analysis of the results can be found in Section~\ref{sec:exp_analysis}. Comparing the F1 scores as well as the OAs obtained on the validation and the test set, respectively, shows that the hyperparameter tuning on the validation set did not result in overfitting as the order of magnitude of the quality metrics on the validation and the test set are en par.
Furthermore, the average F1 scores and the OAs are higher in case of minimizing the focal loss in training. Accordingly, it can be concluded that the classifier is able to better predict the classes of the four tasks by focusing on harder training examples, as is realized in the case of the focal loss. In particular, underrepresented classes benefit more from the use of the focal loss, which is indicated by the larger improvements in terms of the F1 scores compared to the improvements in terms of OA.
The average F1 score over all tasks is 3.7\% higher when using the focal loss compared to minimizing the softmax cross-entropy, whereas the improvement in terms of OA amounts to 0.9\% on average.

\begin{table}[h!]
 \centering
\caption{F1 scores (F1) and overall accuracies (OA) per task (image classification) 
}
\label{tab:MTL_dev_test_eval_object}
\begin{tabular}{lcccccc}
\hline\noalign{\smallskip}
& \multicolumn{2}{c}{train} & \multicolumn{2}{c}{validation} & \multicolumn{2}{c}{test}\\
\hline\noalign{\smallskip}
Variable & F1 & OA & F1  & OA  & F1  & OA  \\
\noalign{\smallskip}\hline\noalign{\smallskip}
\multicolumn{7}{l}{Focal loss} \\
place       &67.3 & 75.4  & 49.2  & 62.5  & 47.0  & 63.1  \\
timespan    &79.3 & 81.9  & 58.4  & 63.8  & 57.5  & 64.5  \\
technique   &87.6 & 88.7  & 75.5  & 79.0  & 77.9  & 80.2  \\
material    &65.4 & 84.8  & 52.2  & 80.6  & 51.2  & 80.6  \\
average     &74.9 & 82.7  & 58.8  & 71.5  & 58.4  & 72.1  \\
\noalign{\smallskip}\hline\noalign{\smallskip}
\multicolumn{7}{l}{Softmax loss} \\
place      &66.7 & 76.3  & 48.2  & 61.0  & 47.2  & 62.2  \\
timespan   &76.8 & 79.8   & 56.0  & 64.4  & 54.2  & 64.9  \\
technique  &84.3 & 85.7  & 72.2  & 75.8  & 74.0  & 76.8  \\
material   &68.6 & 84.6  & 45.0  & 79.4  & 43.4  & 80.7  \\
average    &74.1 & 81.6  & 55.4  & 70.2  & 54.7  & 71.2  \\
\noalign{\smallskip}\hline\noalign{\smallskip}
$\Delta$ (average)   &0.8 & 1.1   & 3.4  & 1.3  & 3.7  & 0.9  \\
\noalign{\smallskip}\hline
\end{tabular}
\end{table}


\subsection{Text Classification}\label{sec:exp:text}
\paragraph{Experimental setup.}
In the text classification experiment we use the method described in Section~\ref{sec:text_classifier}, implemented using PyTorch \cite{NEURIPS2019_9015} and Transformers \cite{wolf-etal-2020-transformers}, and the data described in Section~\ref{sec:data_split} split into training, validation, and test subsets as described in Section~\ref{sec:data_split}. We used the base XLM-R architecture (~125M parameters) with 12-layers, 768-hidden-state, and its respective provided weights. The layers in the classification heads are initialized using the normal distribution $\mathcal{N}(0.0, 0.02)$ with bias parameters set to zero.

First, we performed 50-trial random search hyperparameter tuning implemented using Optuna~\cite{optuna_2019}. During hyperparameter tuning, the text classifier is trained on the train set, and we chose the hyperparameters that resulted in the highest macro F1 score obtained by evaluating on the validation set. These hyperparameters are detailed in Table~\ref{tab:hyper_parameters_text}. We then train a model on the train set with the previously selected hyperparameters and evaluate it on both the validation and test sets. At this stage, the models are allowed to train for up to 20 epochs, but the weights corresponding to the best epoch (on the validation set) are used. This process was performed independently for both the baseline and for each of the methods used for dealing with class imbalance, as described in Section~\ref{sec:balance}. The best parameters refer to the best variant i.e., the one using the focal loss.

\begin{table}[h!]
\centering
\caption{Hyperparameters tuned (text classification)}
\label{tab:hyper_parameters_text}
\begin{tabulary}{\linewidth}{lCc}
\hline\noalign{\smallskip}
Hyperparameter & Range & Best \\
\hline\noalign{\smallskip}
Batch Size $B$          & {4, 8, 32 64} & 64\\
Learning Rate $\eta$    & \{$1e^{-5}$, $2e^{-5}$, 3$e^{-5}$, $5e^{-5}$, $1e^{-4}$ \} & $3e^{-5}$\\
Weight Decay $\omega_R$ & \{0.0, 0.01, 0.02, 0.04 0.05\} & 0.0\\
\noalign{\smallskip}\hline
\end{tabulary}
\end{table}

\paragraph{Results.}
The results of text classification for each strategy dealing with are shown in Table~\ref{tab:text_train_strategy}, with the unweighted softmax loss, trained using proportional sampling, serving as the baseline. The focal loss obtained the best results regarding both the F1 score and $OA$. Random Uniform Sampling also improved the results in terms of F1 score when compared with the baseline, however, the improvement was smaller and came at the cost of some accuracy (0.3\%). Weight Rescaling failed to provide any improvement. 

\begin{table}[h!]
 \centering
\caption{Comparison of strategies for dealing with class imbalance showing average F1 scores (F1) and average overall accuracies (OA) on the validation set (text classification) 
}
\label{tab:text_train_strategy}
\begin{tabular}{lcc}
\hline\noalign{\smallskip}
 & F1  & OA  \\
\noalign{\smallskip}\hline\noalign{\smallskip}
Focal loss (F)              & 86.6  & 89.2    \\
\noalign{\smallskip}\hline\noalign{\smallskip}
Random Uniform Sampling (U) & 85.6  & 88.3 \\
\noalign{\smallskip}\hline\noalign{\smallskip}
Softmax loss (S)            & 85.4  & 88.6    \\
\noalign{\smallskip}\hline\noalign{\smallskip}
Weight Rescaling (W)        & 85.4  & 88.5    \\
\noalign{\smallskip}\hline\noalign{\smallskip}
$\Delta$ (F - U)            & 1.0  & 0.9   \\
$\Delta$ (F - S)            & 1.2  & 0.6   \\
$\Delta$ (F - W)            & 1.2  & 0.7  \\

\noalign{\smallskip}\hline
\end{tabular}
\end{table}

In Table~\ref{tab:mtl_dev_test_eval_text}, we show the results of our best variant, the one using the focal loss, and the standard softmax variant as a baseline. The table presents the accuracy and the average F1 scores, per task, achieved on all records containing text in the validation and test sets. In terms of F1 and overall accuracies, these are the best results for any single modality by a significant margin. This is offset by the fact that the text modality is only present in 39.3\% of all records (Table~\ref{tab:modality_stats}). 

\begin{table}[h!]
  \centering
\caption{F1 scores (F1) and overall accuracies (OA) per task (text classification)}
\label{tab:mtl_dev_test_eval_text}
\begin{tabular}{lcccccc}
\hline\noalign{\smallskip}
&\multicolumn{2}{c}{train} &\multicolumn{2}{c}{validation} & \multicolumn{2}{c}{test}\\
\hline\noalign{\smallskip}
Variable  & F1  & OA & F1  & OA  & F1  & OA  \\
\noalign{\smallskip}\hline\noalign{\smallskip}
\multicolumn{7}{l}{Focal loss (F)} \\  
place       &99.9 & 99.9 & 92.6 & 91.8 & 93.9 & 93.3 \\
timespan    &99.9 & 99.9 & 85.7 & 90.5 & 84.3 & 89.5 \\
technique   &99.5 & 99.5 & 86.8 & 88.0 & 87.8 & 89.5 \\
material    &99.0 & 99.2 & 81.4 & 86.3 & 78.2 & 84.6 \\
average     &99.6 & 99.6 & 86.6 & 89.2 & 86.1 & 89.2 \\
\noalign{\smallskip}\hline\noalign{\smallskip}
\multicolumn{7}{l}{Softmax loss (S)} \\  
place       &99.6 & 99.7 & 92.2 & 91.3 & 93.1 & 92.6 \\
timespan    &98.8 & 99.4 & 83.8 & 89.8 & 82.1 & 88.0 \\
technique   &97.3 & 97.7 & 87.1 & 88.2 & 88.0 & 89.7 \\
material    &97.0 & 97.5 & 78.5 & 85.0 & 78.7 & 85.4 \\
average     &98.2 & 98.6 & 85.4 & 88.6 & 85.5 & 88.9 \\
\noalign{\smallskip}\hline\noalign{\smallskip}
$\Delta$ (average)   &1.4 & 1.0 & 1.2  & 0.6  & 0.6  & 0.3  \\
\noalign{\smallskip}\hline
\end{tabular}
\end{table}

The improvement provided by the use of the focal loss, in terms of F1, is lower on the test set when compared to the validation set (0.6\% vs 1.2\%). When the results are broken down by variable, we see that the focal loss obtained worse results in \textit{technique} and \textit{material}. Both tasks have particular challenges, as discussed in Section~\ref{sec:analysis:text}. To summarize here, they are expected to have more noisy labels, which in turn may interact negatively with the focal loss, since the "harder" examples the loss focuses on are more likely to be label noise. In comparison with image classification, incorrect labels are much more evident from the text descriptions (see Tables~\ref{tab:text_examples} and~\ref{tab:misleading_text_examples}) than from an image, as direct or subclass references are often present in the text. This effect is probably compounded by the smaller number of records with text (Table~\ref{tab:class_stats}) and the relatively higher accuracy of the text classifier.

\subsection{Tabular Classification}
\label{sec:exp:tab}
\paragraph{Experimental setup.}
For each task, we train an individual classifier with different parameters and hyperparameters, selected by task-specific hyperparameter tuning using grid search. We show the hyperparameters, the search space for tuning, and the selected values in Table~\ref{tab:hyper_parameters_tab}. Note that the ranges used were all within very reasonable intervals as an additional guard against overfitting. Weight rescaling was directly incorporated into parameter tuning by providing either no sample weights or balanced sample weights to the grid search. Since we have independent classifiers per task, it's entirely sensible to use weight rescaling only for certain tasks.

\begin{table*}[h!]
\centering
\caption{Hyperparameters tuned (tabular classification)}
\label{tab:hyper_parameters_tab}
\begin{tabular}{lcccccc}
\hline\noalign{\smallskip}
Hyperparameter         & Range         & Interval  & place & timespan & technique & material \\
\hline\noalign{\smallskip}
colsample{\_}bytree     & [0.6, 1.0]    & 0.2       & 0.8    & 0.8     & 0.6       & 0.8 \\
gamma                   & [0.0, 0.4]    & 0.2       & 0.4    & 0.2     & 0.2       & 0.0 \\
learning{\_}rate        & [0.1, 0.3]    & 0.1       & 0.3    & 0.3     & 0.3       & 0.2 \\
max{\_}depth            & [2, 8]        & 2         & 4      & 4       & 4         & 8 \\
min{\_}child{\_}weight  & [1, 4]        & 1         & 1      & 2       & 4         & 2 \\
n{\_}round              & [100, 500]    & 400       & 100    & 100     & 100       & 500 \\
subsample               & [0.6, 1.0]    & 0.2       & 0.6    & 1.0     & 0.6       & 0.8 \\
sample weight           & \{"none", "balanced"\}&-  & "none" & "none"  & "none"    & "none" \\
\noalign{\smallskip}\hline
\end{tabular}
\end{table*}

\paragraph{Results.}
We show the evaluation results in Table~\ref{tab:tab_dev_test_eval}. Given that it essentially relies on co-occurrences of very coarse labels, the results seem reasonable.
In fact, in terms of F1 and accuracy, they almost match the image classifier. Within the hyperparameter tuning, weight rescaling was not selected for any task.

\begin{table}[h!]
\centering
\caption{F1 (F1) and overall accuracies (OA) obtained in the experiment per task (tabular classification)}
\label{tab:tab_dev_test_eval}
\begin{tabular}{lcccccc}
\hline\noalign{\smallskip}
& \multicolumn{2}{c}{train}& \multicolumn{2}{c}{validation} & \multicolumn{2}{c}{test}\\
\hline\noalign{\smallskip}
Variable   & F1  & OA & F1  & OA & F1 & OA \\
\noalign{\smallskip}\hline\noalign{\smallskip}      
place       & 49.0 & 63.0 & 47.9  & 62.4 & 46.2 & 61.9 \\
timespan    & 60.4 & 67.6 & 57.5 & 65.1 & 58.6  & 67.6 \\
technique   & 69.1 & 73.8 & 68.6 & 74.2 & 68.3 & 73.0 \\
material    & 54.4 & 83.2 & 50.7 & 82.1 & 49.4 & 82.1  \\
\noalign{\smallskip}\hline\noalign{\smallskip}
average     & 58.2 & 71.9 & 56.2 & 71.0 & 55.6 & 71.2 \\
\noalign{\smallskip}\hline
\end{tabular}
\end{table}

Feature importance by gain is shown in Table \ref{tab:tab_feature_importance}. 
For every task, the tabular classifier's most important feature is the museum. 
That could probably be expected because museums are not random collections of objects. 

\begin{table}[h!]
\centering
\caption{Tabular classifier feature importance per task (information gain)}
\label{tab:tab_feature_importance}
\setlength\tabcolsep{2pt}
\begin{tabular}{lccccc}
\hline\noalign{\smallskip}
Target      & \multicolumn{5}{c}{Feature} \\
Variable    & museum & place & timespan & technique & material \\
\noalign{\smallskip}\hline\noalign{\smallskip}
place       &  \textbf{0.49} & -     &  0.20    & 0.12      & 0.19 \\
timespan    &  \textbf{0.41} & 0.31  & -        & 0.16      & 0.12 \\
technique   &  \textbf{0.40} & 0.29  & 0.17     & -         & 0.14 \\
material    &  \textbf{0.39} & 0.21  & 0.16     & 0.24      & -  \\
\noalign{\smallskip}\hline
\end{tabular}
\end{table}

\subsection{Multimodal Classification}
\label{sec:exp:multimodal}
\paragraph{Experimental setup.}
For the experiments involving multimodal classifiers, we started by training the three classifiers based on single modalities (images, text, tabular, respectively)  on the training set independently of each other in the way described in Sections~\ref{sec:exp:image}~-~\ref{sec:exp:tab}. 
After that, these classifiers were used to classify the samples in the validation set.
Finally, we used these predictions as inputs to train the multimodal classifier on the validation set. 
We used five-fold cross-validation on the validation set to perform hyperparameter tuning using grid search for the same set of hyperparameters that were used in tuning the tabular classifier. The ranges of the hyperparameters are slightly scaled down to reflect the smaller size of the training data. The details of hyperparameter tuning are shown in Table~\ref{tab:hyper_parameters_multimodal}. 

\begin{table*}[h!]
  \centering
\caption{Hyperparameter tuning (multimodal classifier)}
\label{tab:hyper_parameters_multimodal}
\begin{tabular}{lcccccc}
\hline\noalign{\smallskip}
Hyperparameter         & Range         & Interval  & place & timespan & technique & material \\
\hline\noalign{\smallskip}
colsample{\_}bytree     & [0.6, 1.0]    & 0.2       & 0.6    & 1.0     & 0.8       & 1.0 \\
gamma                   & [0.0, 0.4]    & 0.2       & 0.4    & 0.2     & 0.4       & 0.4 \\
learning{\_}rate        & [0.1, 0.2]    & 0.1       & 0.1    & 0.1     & 0.1       & 0.1 \\
max{\_}depth            & [2, 6]        & 2         & 6      & 2       & 6         & 2 \\
min{\_}child{\_}weight  & [1, 2]        & 1         & 2      & 1       & 2         & 1 \\
n{\_}round              & [100, 300]    & 100       & 300    & 100     & 200       & 100 \\
subsample               & [0.6, 1.0]    & 0.2       & 1.0    & 1.0     & 0.6       & 0.8 \\
sample weight           & \{"none", "balanced"\}&-  & "none. & "none"  & "none"    & "none" \\
\noalign{\smallskip}\hline
\end{tabular}
\end{table*}

\paragraph{Results.}
The results of the experiments are shown in Table \ref{tab:multimodal_eval}. As in the tabular classifier, weight rescaling using the "balanced" heuristic failed to improve results.
\begin{table}[h!]
\centering
\caption{F1 scores (F1) and overall accuracies (OA) per task (multimodal classifier)}
\label{tab:multimodal_eval}
\begin{tabular}{lcc}
\hline\noalign{\smallskip}
Variable & F1  & OA \\
\noalign{\smallskip}\hline\noalign{\smallskip}        
place       & 77.6  & 79.6  \\
timespan    & 74.2  & 80.2  \\
technique   & 83.6  & 85.0   \\
material    & 61.3  & 85.5   \\
\noalign{\smallskip}\hline\noalign{\smallskip}
average     & 74.2  & 82.6   \\
\noalign{\smallskip}\hline

\end{tabular}
\end{table}

The overall accuracy and mean F1 score achieved by the multimodal classifier are better than those for the image classifier (Table~\ref{tab:MTL_dev_test_eval_object}) and slightly worse than those reported for the text classifier (Table~\ref{tab:mtl_dev_test_eval_text}), but this comparison is inconclusive because the results in Table~\ref{tab:mtl_dev_test_eval_text} only consider records for which text is available, which is only about 39\% of the test set, whereas the evaluation of the image classifier is based on about 96\% of the test set and multimodal classification is based on the complete test set. 

For the majority of the samples, only images and tabular information are available, and thus the prediction would be based on these modalities. 
In order to compare the results of all modalities, we carried out an evaluation of all modality-specific classifiers and the multimodal classifier on the entire test set. 
In this evaluation, a record for which a modality is missing was considered a wrong prediction for that modality-specific classifier. 
For instance, a record without images was considered to be an incorrect prediction for the image classifier. The results of this evaluation are shown in Table~\ref{tab:comparison}. In this comparison, the multimodal classifier very significantly outperforms all single modality classifiers (74.2\% vs 55.6\%). 

\begin{table}[h!]
\centering
\caption{F1 scores evaluated on the entire test set (all classifiers)}
\label{tab:comparison}
\begin{tabular}{lcccc}
\hline\noalign{\smallskip}
Variable  & image & text  &tabular   & multimodal\\ 
\noalign{\smallskip}\hline\noalign{\smallskip}
place       & 38.0               & 65.0               & 46.2                   & \textbf{77.6}\\
timespan    & 49.2               & 55.6              & 58.6                   & \textbf{74.2} \\
technique   & 73.5               & 41.0               & 68.3                  & \textbf{83.6}\\
material    & 46.5               & 37.4              & 49.4                   & \textbf{61.3}\\
\noalign{\smallskip}\hline\noalign{\smallskip}
average     & 51.8               & 49.8               & 55.6                   & \textbf{74.2}\\
\noalign{\smallskip}\hline
\end{tabular}
\end{table}

We also performed an ablation study to assess the importance of the individual modalities for the classification results 
The ablation study was performed by removing one of the modalities from the input of the fusion classifier, leaving only the other two modalities (Table~\ref{tab:multimodal_eval_mod_comb}). The results of the modality ablation study and the individual modality results (Table~\ref{tab:comparison}), confirm the assumption that each modality provides a meaningful contribution. The results of a multimodal classifier that combines any two modalities are superior to those of any individual modality. Further, combining all three produces the best results.

\begin{table}[h!]
\centering
\caption{Average F1 scores of the multimodal classifier using different input modalities (average over all tasks)}
\label{tab:multimodal_eval_mod_comb}
\begin{tabular}{lc}
\hline\noalign{\smallskip}
Input Modality  &  F1 \\
\noalign{\smallskip}\hline\noalign{\smallskip}
text + tabular  & 71.6  \\
image + text    & 70.6\\
image + tabular & 62.4 \\
\noalign{\smallskip}\hline\noalign{\smallskip}
all     & \textbf{74.2} \\
\noalign{\smallskip}\hline
\end{tabular}
\end{table}


\section{Analysis}
\label{sec:exp_analysis}

\subsection{Image classification}
Here, we will provide a more detailed analysis of the results of the CNN-based image classifier in Table~\ref{tab:MTL_dev_test_eval_object}. The table shows that the classification performance strongly varies between tasks. Comparing the OAs, one can see that the variable \textit{material} achieves the highest OAs, followed by \textit{technique} and \textit{timespan}; the worst OA is achieved for the variable \textit{place}. Taking the class structure shown in Table~\ref{tab:class_stats} into account, a connection can be made to the number of classes constituting a task's class structure. The larger the number of classes to be distinguished, the lower the achieved percentage of correctly classified images in the softmax experiment, where a similar behavior can be observed for the focal experiment;
\textit{material} having three classes has the highest OA of 80.7\%, followed by \textit{technique} having four classes with 76.8\% correct predictions and \textit{timespan} with five classes with a OA of 64.0\%, whereas \textit{place} with nine classes has the lowest OA of 62.2\%.

An analysis of the task-specific F1 scores in connection with the class distributions of the respective task indicates a dependency of the F1 score on the degree of class imbalance.
Taking the ratio of the number of image examples for the majority class, i.e., the class with the most labeled examples in the dataset, in relation to the number of image examples for the minority class, i.e., the class with the fewest examples, a negative correlation between this ratio and the achieved task-specific F1 score can be observed for the focal loss experiment, where a similar behavior can be observed for the softmax experiment.
The majority class of \textit{technique} has 2.5 times as many examples as the minority class and \textit{technique} has the highest F1 score of 77.9\%, followed by \textit{timespan} with a ratio of 4.9 and a score of 57.5\% and \textit{material} with a ratio of 7.7 having a score of 51.2\%. The lowest F1 score of 47.0\% is obtained for \textit{place} with a ratio of 8.3. We attempted to overcome this dependency of the F1 scores on the class distributions by focusing on hard training examples through the presented variant of the focal loss in equation \ref{eq:loss_incomp_MTL_focal}.
Analyzing the improvements of the F1 scores by utilizing the focal loss instead of the softmax cross-entropy loss shows that the focal loss indeed reduces this dependency. Except for the variable \textit{place}, there is an improvement of the task-specific F1 scores, and in these cases, it is larger for tasks with higher class imbalance (indicated by a high ratio between the number of examples for the majority class and the minority class, respectively). The F1 score of \textit{material} (ratio of 7.7) is improved by 7.8\%, whereas the F1 score of \textit{technique} (ratio of 2.5) is improved by 3.9\%.
The variable \textit{place} with a ratio of 8.3 should have received the largest improvement in F1 score according to the general trend, but it actually is slightly worse (-0.2\%). 
We assume this to be related to the large number of classes to be distinguished for \textit{place}, which might make a correct prediction more complicated for this variable than for the other ones.

\begin{figure}[h]
     \centering
     \begin{subfigure}[b]{0.25\textwidth}
         \centering
         \includegraphics[width=\textwidth]{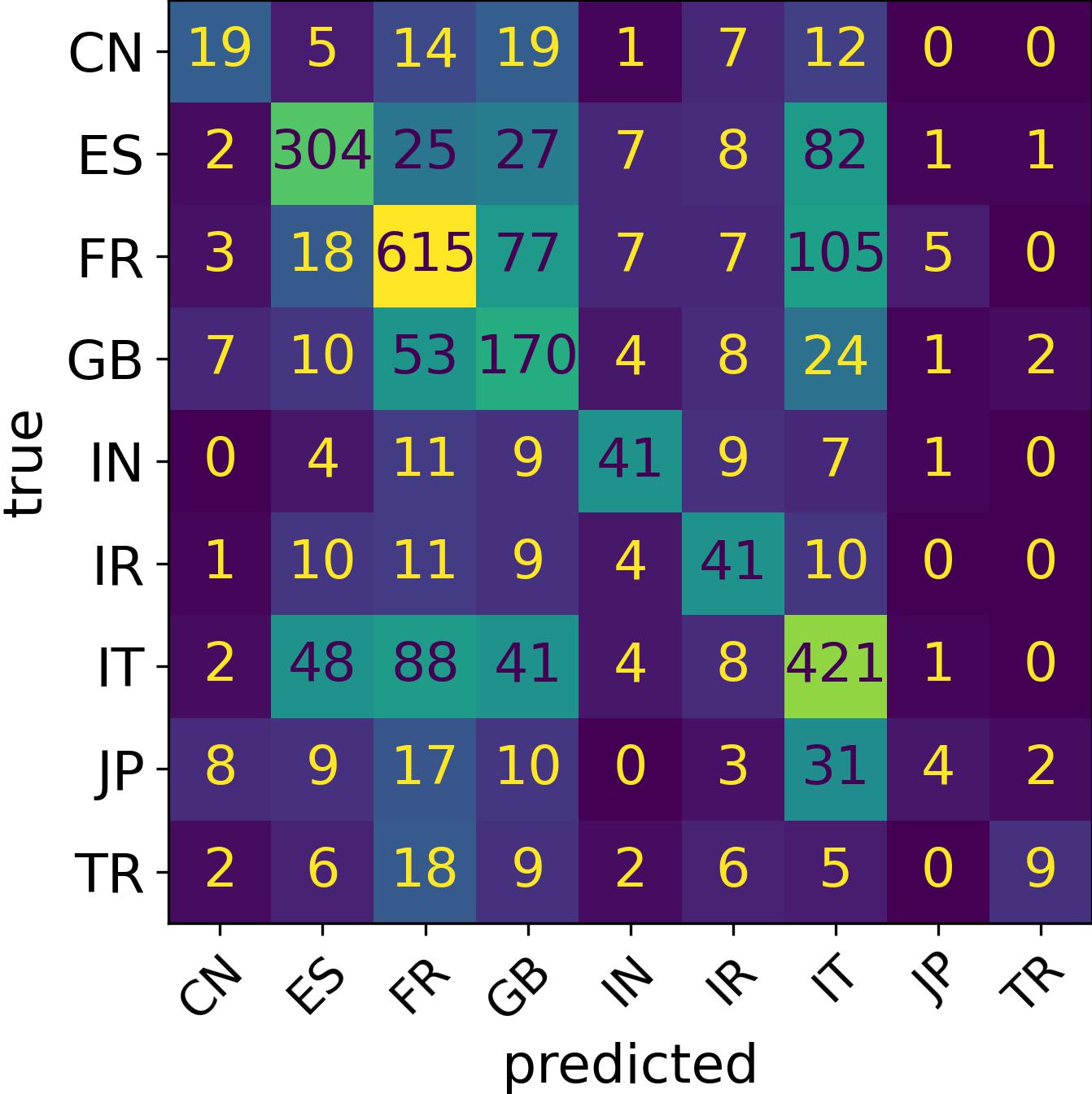}
         \caption{place}
         \label{fig:img:cm:place}
     \end{subfigure}
    \hfill
     \begin{subfigure}[b]{0.21\textwidth}
         \centering
         \includegraphics[width=\textwidth]{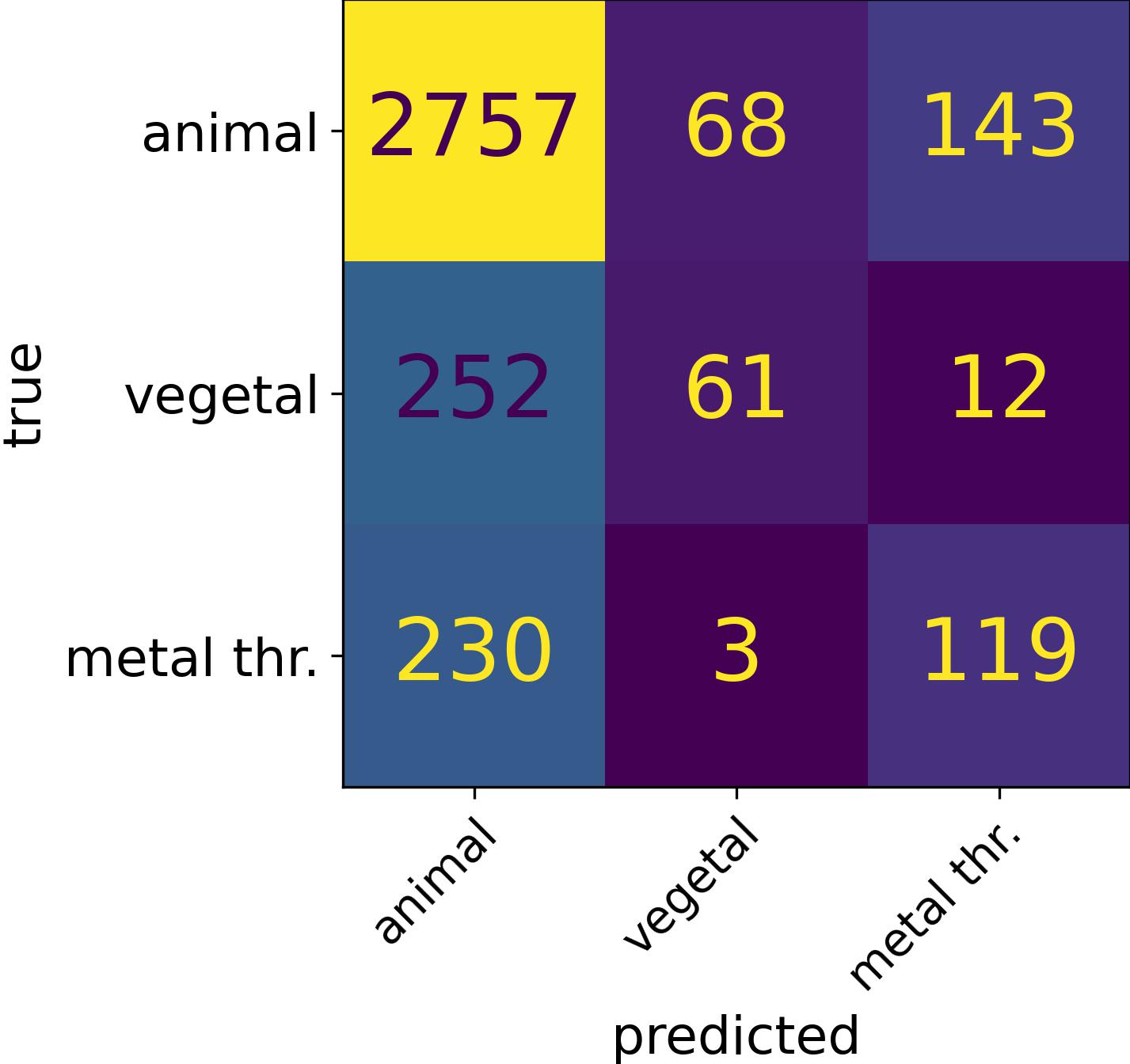}
         \caption{material}
         \label{fig:txt:img:material}
     \end{subfigure}
     \hfill
     \begin{subfigure}[b]{0.25\textwidth}
         \centering
         \includegraphics[width=\textwidth]{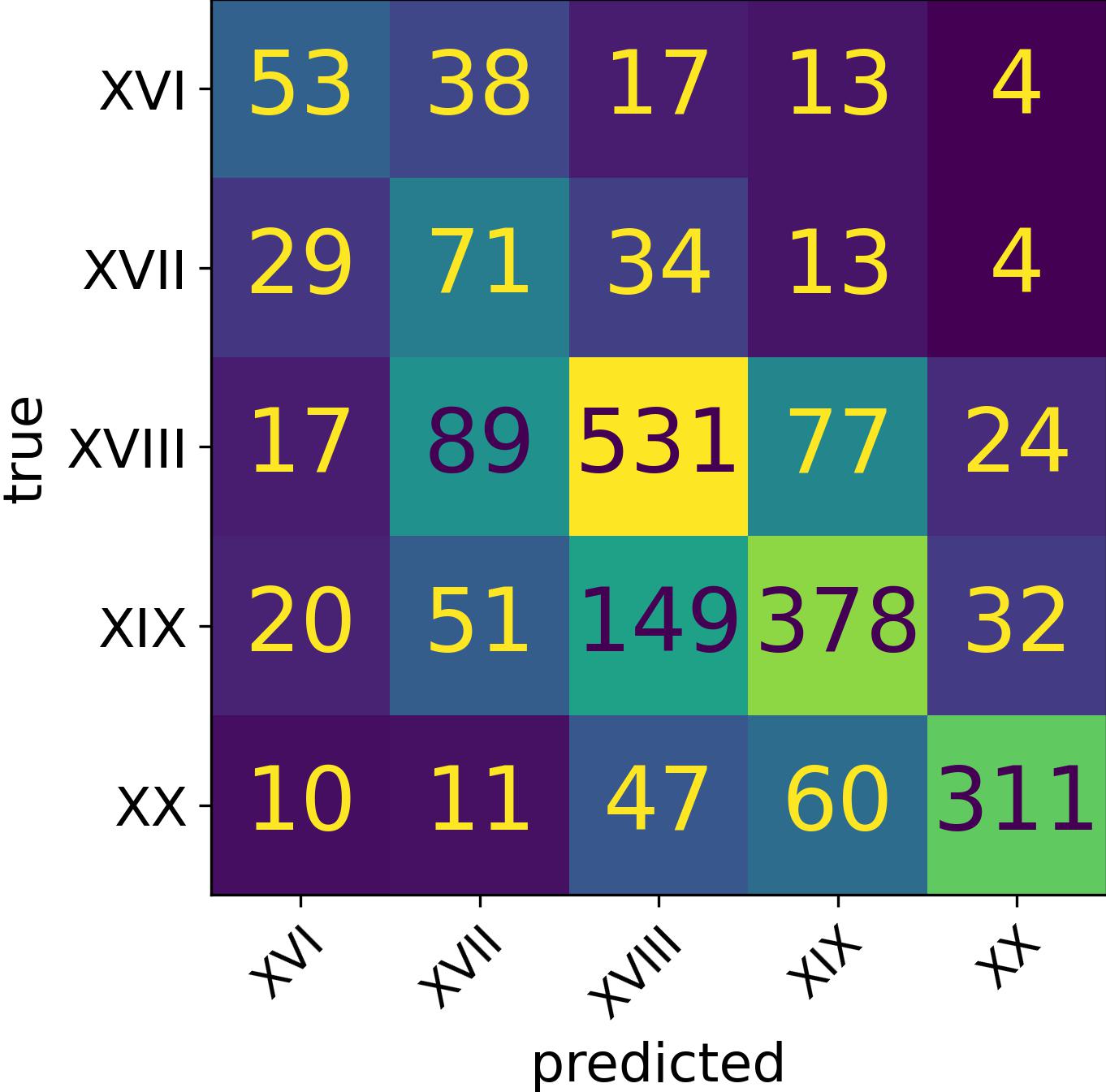}
         \caption{timespan}
         \label{fig:img:cm:timespan}
     \end{subfigure}
     \hfill
     \begin{subfigure}[b]{0.21\textwidth}
         \centering
         \includegraphics[width=\textwidth]{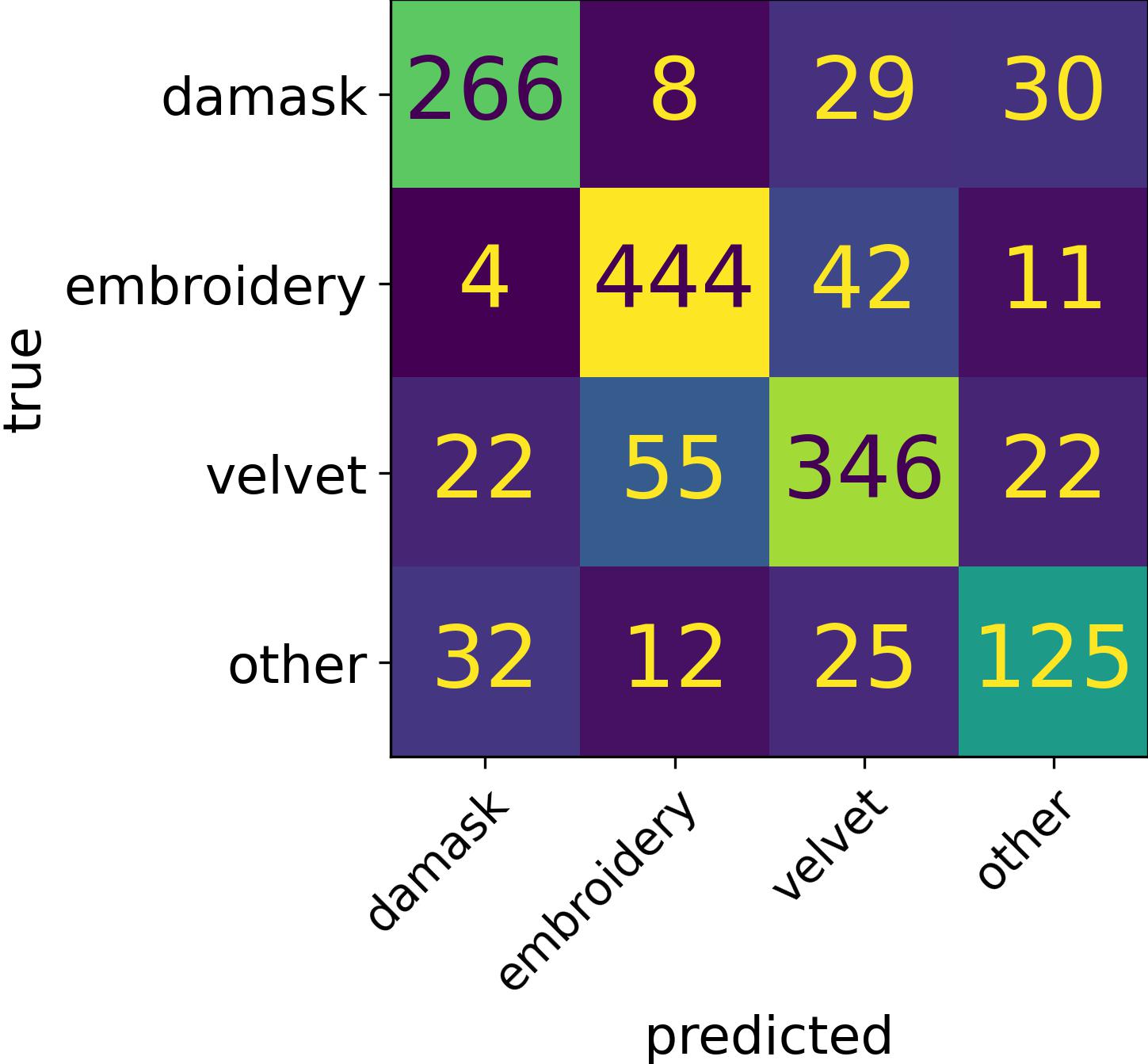}
         \caption{technique}
         \label{fig:img:cm:technique}
     \end{subfigure}

    \caption{Image classifier confusion matrices}
    \label{fig:img:cm}
\end{figure}

In summary, the utilization of the focal loss improves the performance of the trained classifier in correctly predicting the properties of silk based on images.
Even though the variable-specific F1 score still seems to depend on the degree of imbalance of a task's class distribution, as can be seen in Figure~\ref{fig:img:cm}, focusing on hard examples during training primarily improves the task-specific F1 scores of tasks with large class imbalances, as long as the number of classes to be differentiated is not too large.
Solving the remaining challenge of predicting all classes of a task equally well may require more data, as not all aspects of all silk properties are equally well represented in the available images.

\subsection{Text classification}
\label{sec:analysis:text}
The results for the text modality, shown in Table~\ref{tab:mtl_dev_test_eval_text}, are better, in terms of F1 and OA, than the results for the image modality (Table~\ref{tab:MTL_dev_test_eval_object}) or the tabular modality (Table~\ref{tab:tab_dev_test_eval}). This is not surprising, since the properties we are predicting are important to domain experts. They, or their taxonomy subclasses, are often included in the text descriptions of the cultural heritage objects (although not necessarily using the same words). Even when they are not, we can intuitively expect some degree of similarity between text descriptions of objects with similar underlying values for these properties, either globally or at least within the same museum.

The biggest disadvantage of the text classifier is that text descriptions are present in the dataset far less often than images, as shown in Table~\ref{tab:modality_stats}. Once adjusted for missing modality, given that more than 60\% of the records are missing this modality, the text classifier actually performs the worst of any modality, as shown in Table~\ref{tab:comparison}.

We analyzed about 20 misclassified English language test set examples for each task. In around half of the cases, there was no direct information that could've allowed an accurate classification. E.g., no location is mentioned when attempting to classify place, or no date is mentioned when attempting to classify timespan. This forces the classifier to rely on other statistical regularities present in the text to provide a classification.

The \textit{material} task is particular. Its most common class, "animal fibre", is a de facto background class. All records in the dataset should be of silk fabrics, which means the material they are made of is an "animal fibre". Some have other materials too. These other materials can correspond to a vegetable fibre (e.g., cotton) or a thread with some metal (e.g., gold thread). While the problem of not having label-specific information in the text is common in the examples we analyzed (6/20), obviously incorrectly labeled examples were even more common (9/20). This occurs when either the original record is missing the correct label or when the automatic extraction and linking of the label fails. The high prevalence of this type of error within this task in the examples we analyzed, combined with its absence in other tasks, leads us to suspect that this is the main cause of the relatively lower accuracy and F1 scores for this task.

The \textit{technique} task is also particular in terms of the examples we analyzed. For one, it is the only variable with a catch-all class, "other" that contains examples of objects primarily produced using techniques other than those with an explicit class label. Additionally, a significant number of examples (5/19) contain information that would imply multiple labels, where, usually, a small part of the object was produced using a different technique from the main part of the object. A similar type of error occurs in the \textit{timespan} task within a similar proportion of examples (5/10). In the \textit{timespan} task, this can occur when an object was produced at a certain date but later altered or when the estimated date of production within the text crossed centuries. 
We believe the properties of \textit{material} and \textit{technique} likely resulted in more label noise, particularly evident from the text descriptions. We hypothesize that this noise might have undesirable interactions with the focal loss that might explain its worse results in these tasks. However, further study is necessary to draw firm conclusions.

We hypothesize that the somewhat better results for the place task are connected to regularities between the museum and an object's place of production. This connection is suggested in Table~\ref{tab:tab_feature_importance}. Text descriptions are very indicative of the museum, not just in the language but usually also in style, length, and topics.

We would like to point out the existence of misleading text examples such as in Table~\ref{tab:misleading_text_examples} where information in the text can correspond to an incorrect label. We do this to give the reader a better understanding of the challenges faced by an algorithm.

\begin{table*}[h]
\centering
\caption{Examples of misleading text descriptions with emphasis added to highlight the misleading text snippets}
\label{tab:misleading_text_examples}
\begin{tabulary}{\linewidth}{Lcc}
    \hline
Text Description (Snippet)    & Predicted & True                                           \\
    \hline
    {\scriptsize motifs found on \textbf{seventeenth-century} coverlets but must have been made in the early eighteenth century (...) Embroidery of Gujerat [sic] in the \textbf{Seventeenth Century}}  & XVII & XVIII \\
    \hline
     {\scriptsize derived from engravings after Maarten de Vos which first appeared in Gerard de Jode's \textbf{1579} illustrated bible}  & XVI & XVII \\
    \hline
     {\scriptsize Center text reads \textbf{"Vole vole mon coeur!}} & FR & GB \\
     \hline
     {\scriptsize Depiction from the \textbf{Italian} poem} & IT & GB \\
     \hline
\end{tabulary}
\end{table*}

\begin{figure}[h]
     \centering
     \begin{subfigure}[b]{0.25\textwidth}
         \centering
         \includegraphics[width=\textwidth]{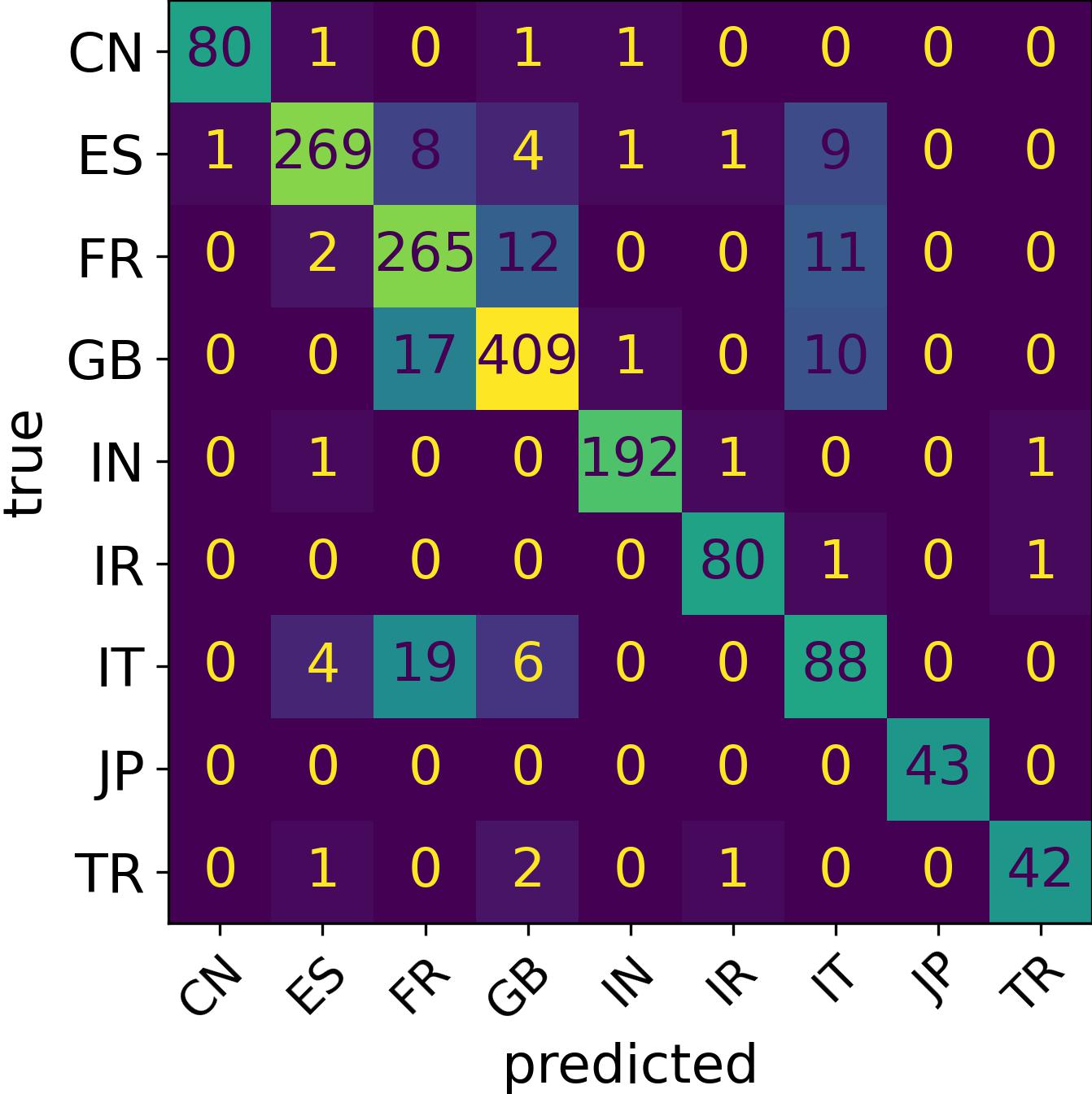}
         \caption{place}
         \label{fig:txt:cm:place}
     \end{subfigure}
    \hfill
     \begin{subfigure}[b]{0.21\textwidth}
         \centering
         \includegraphics[width=\textwidth]{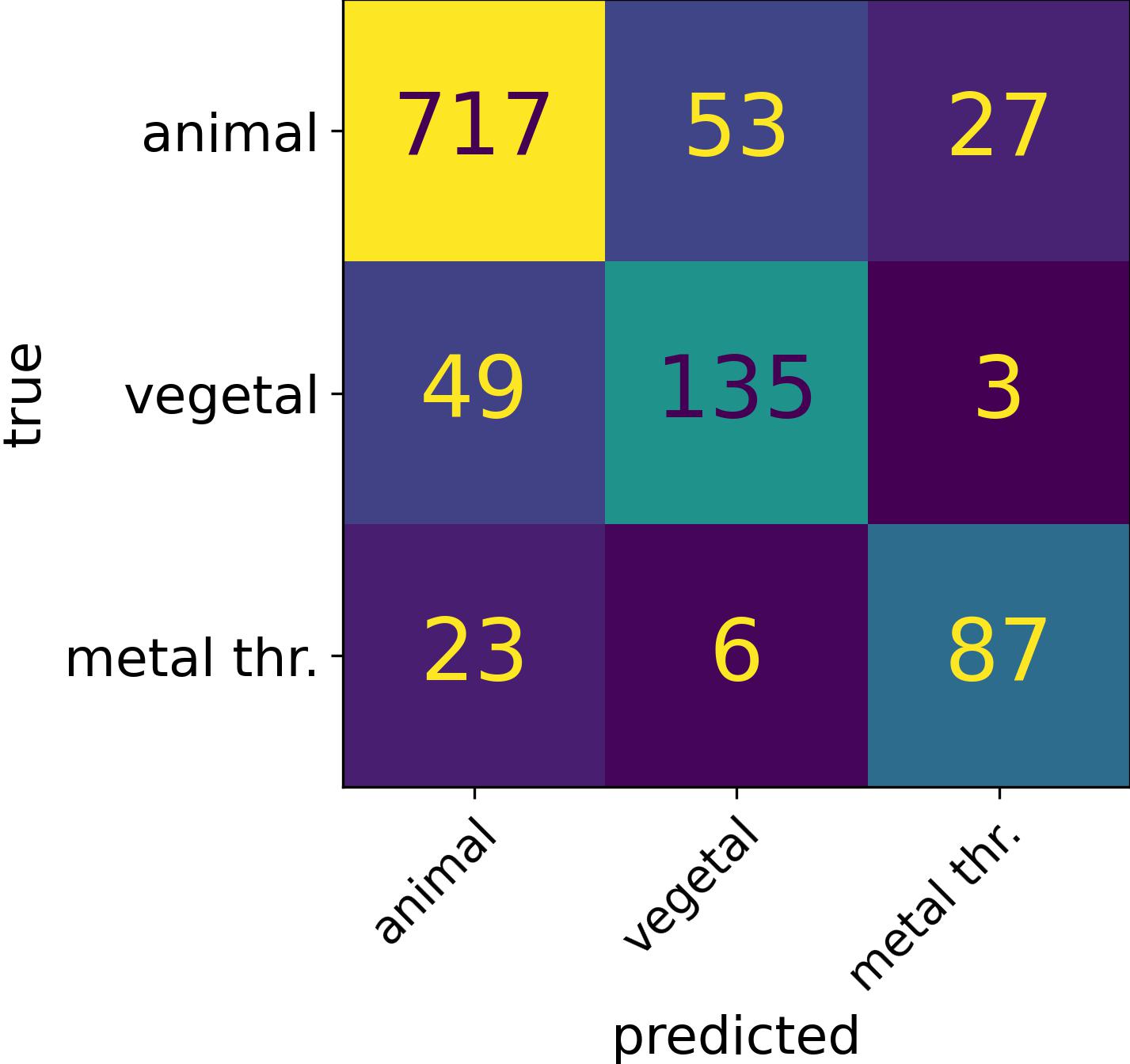}
         \caption{material}
         \label{fig:txt:cm:material}
     \end{subfigure}
     \hfill
     \begin{subfigure}[b]{0.25\textwidth}
         \centering
         \includegraphics[width=\textwidth]{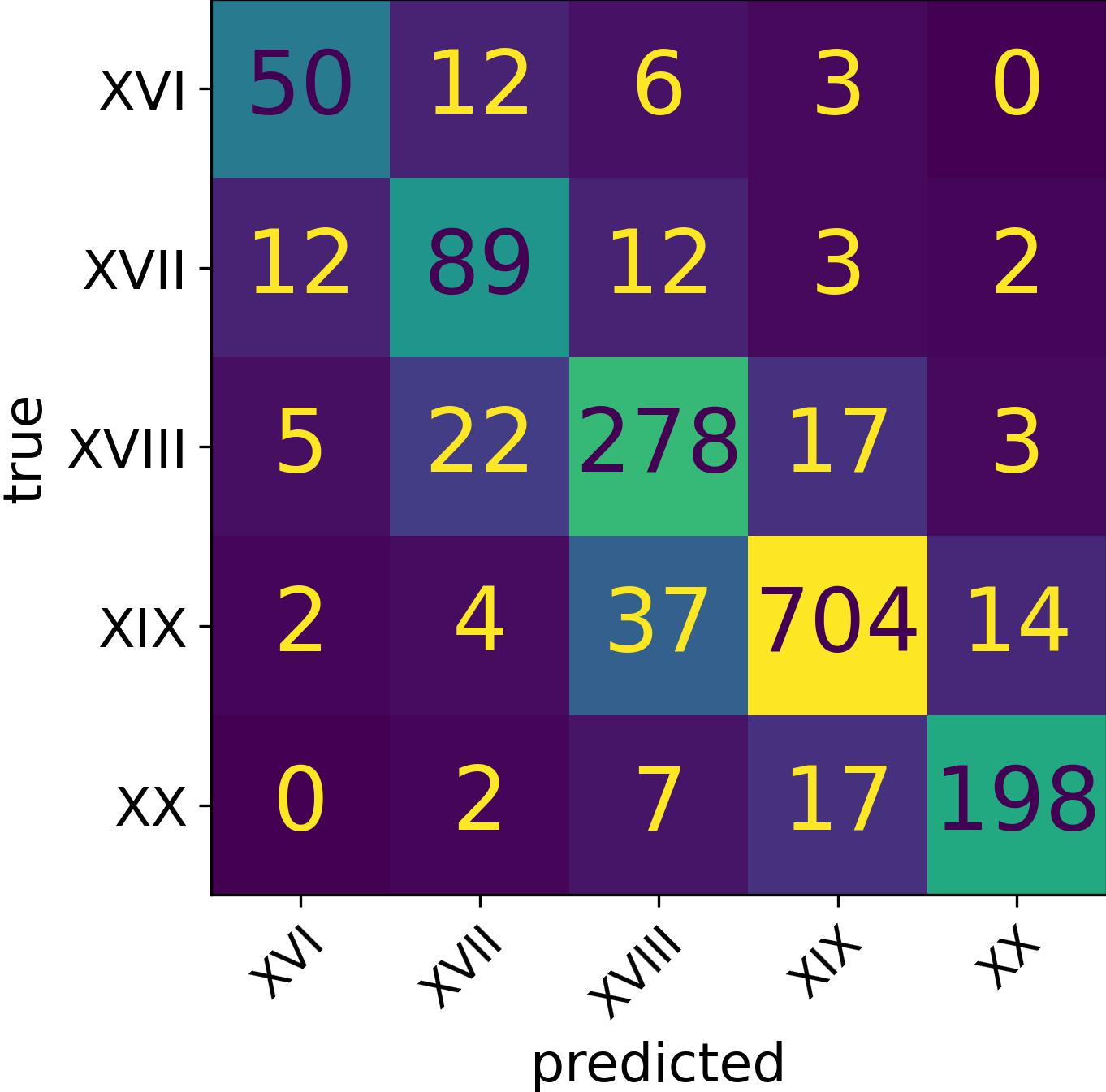}
         \caption{timespan}
         \label{fig:txt:cm:timespan}
     \end{subfigure}
     \hfill
     \begin{subfigure}[b]{0.21\textwidth}
         \centering
         \includegraphics[width=\textwidth]{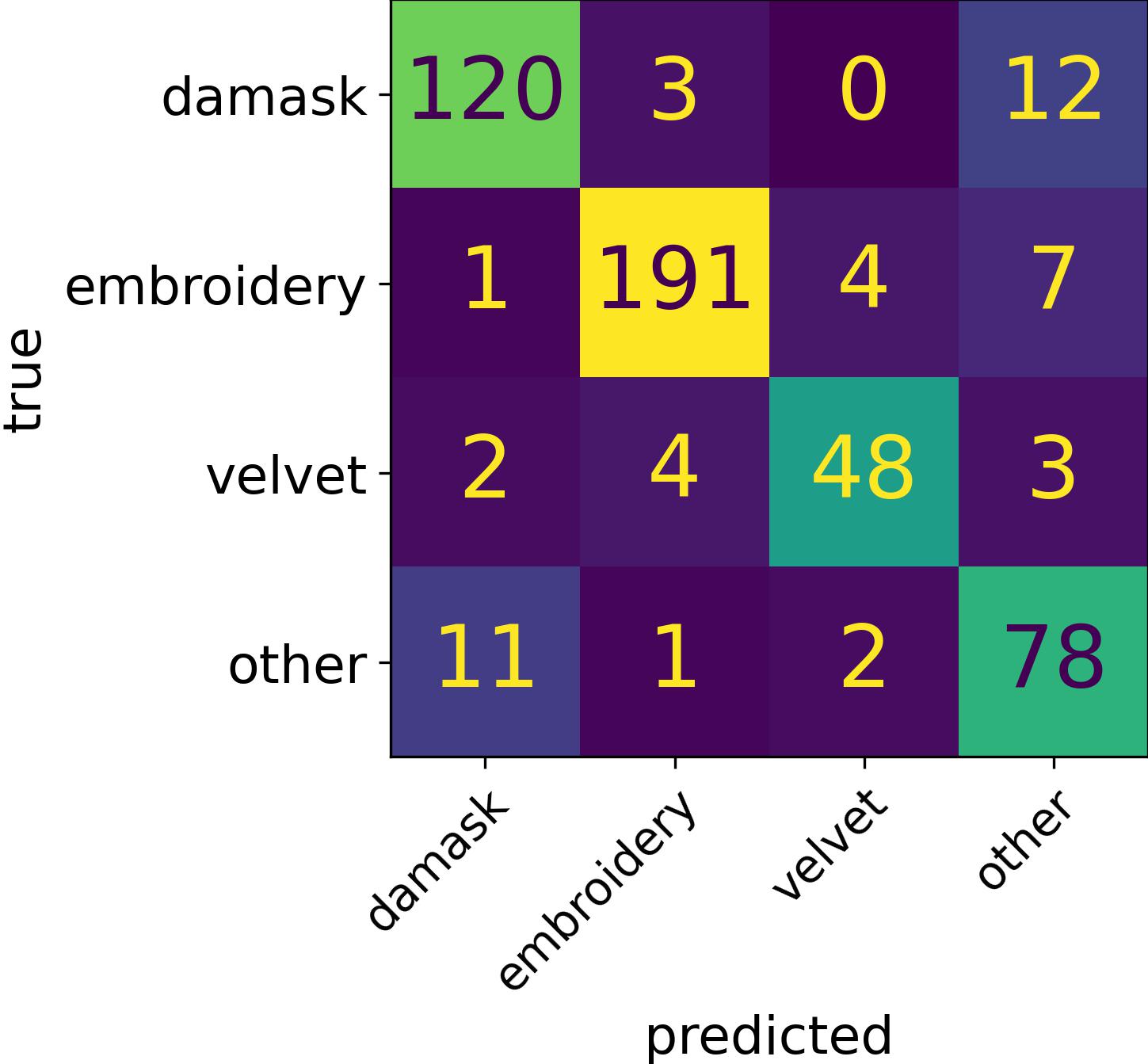}
         \caption{technique}
         \label{fig:txt:cm:technique}
     \end{subfigure}
    \caption{Text classifier confusion matrices}
    \label{fig:text:cm}
\end{figure}

Figure~\ref{fig:text:cm} shows the confusion matrices for text classification. Errors for the place property seem to be primarily between near European countries labels (IT-ES-FR-GB). These are also the most common classes. In the timespan task, errors are clearly clustered around the diagonal, indicating errors mostly between chronologically near dates. For material, most errors represent false positives for "animal fibre", followed closely by "vegetable fibre". Most errors in the technique task seem to be false "other". This may be due to label noise. The text classifier is the exceptional modality in terms of the errors it makes. Not only does it make fewer errors, but these are often qualitatively different. In essence, text carries more easily accessible relevant information for these tasks. On the flip side, it's probably in the unique position of having to deal with misleading examples.

\subsection{Tabular classification}
\begin{figure}[h]
     \centering
         \includegraphics[width=0.3\textwidth]{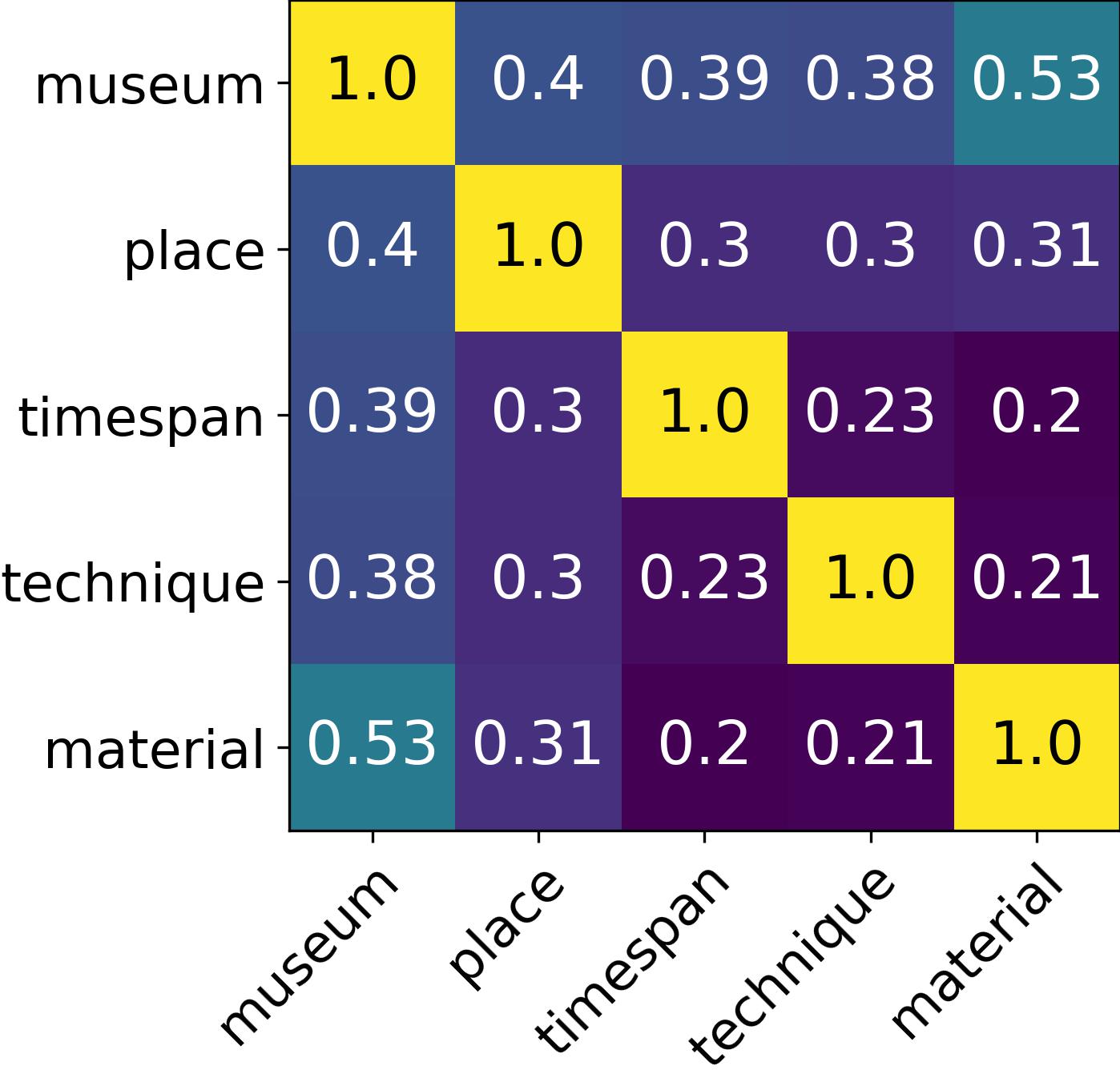}
         \caption{Association between features of the tabular classifier (Cramer's V).}
         \label{fig:tab:cramers}
\end{figure}

When compared to the other modalities, the tabular classifier performs (Table~\ref{tab:tab_dev_test_eval}) slightly worse than the image classifier (Table~\ref{tab:MTL_dev_test_eval_object}) in terms of average F1 score, respectively, 55.6\% versus 58.4\%. However, this situation is reversed when accounting for missing modality (Table~\ref{tab:comparison}), with the tabular classifier outperforming the image classifier with F1 scores of 55.6\% versus 51.8\%. Intuitively, from a domain perspective, we can expect these variables to be associated. For example, a certain country is more active in producing textiles during a certain timespan than during others. Furthermore, museums are typically curated and not random collections. However, given the limited number of features and the coarseness of the labels, we should not overestimate the strength of the association between variables, which we calculated as Cramer's V in Figure~\ref{fig:tab:cramers}. Cramers' V is a symmetric measure that gives a value between 0 and 1 for the association between two nominal variables. We see that the values are, by themselves, relatively low, with only \textit{material} and \textit{museum} having a value above 0.5. The association between \textit{museum} and the other properties was the highest (first column or row) which again reinforces the belief that the curated nature of museum collections and its impact on the other properties is learnable from this dataset.
The strength of the association does not directly translate into feature importance as measured by information gain (Table~\ref{tab:tab_feature_importance}). Rather, it seems that a particular combination of these associations is being learned by the tree boosting algorithm in such a way that results in a relatively effective classifier.

\begin{figure}[h]
     \centering
     \begin{subfigure}[b]{0.25\textwidth}
         \centering
         \includegraphics[width=\textwidth]{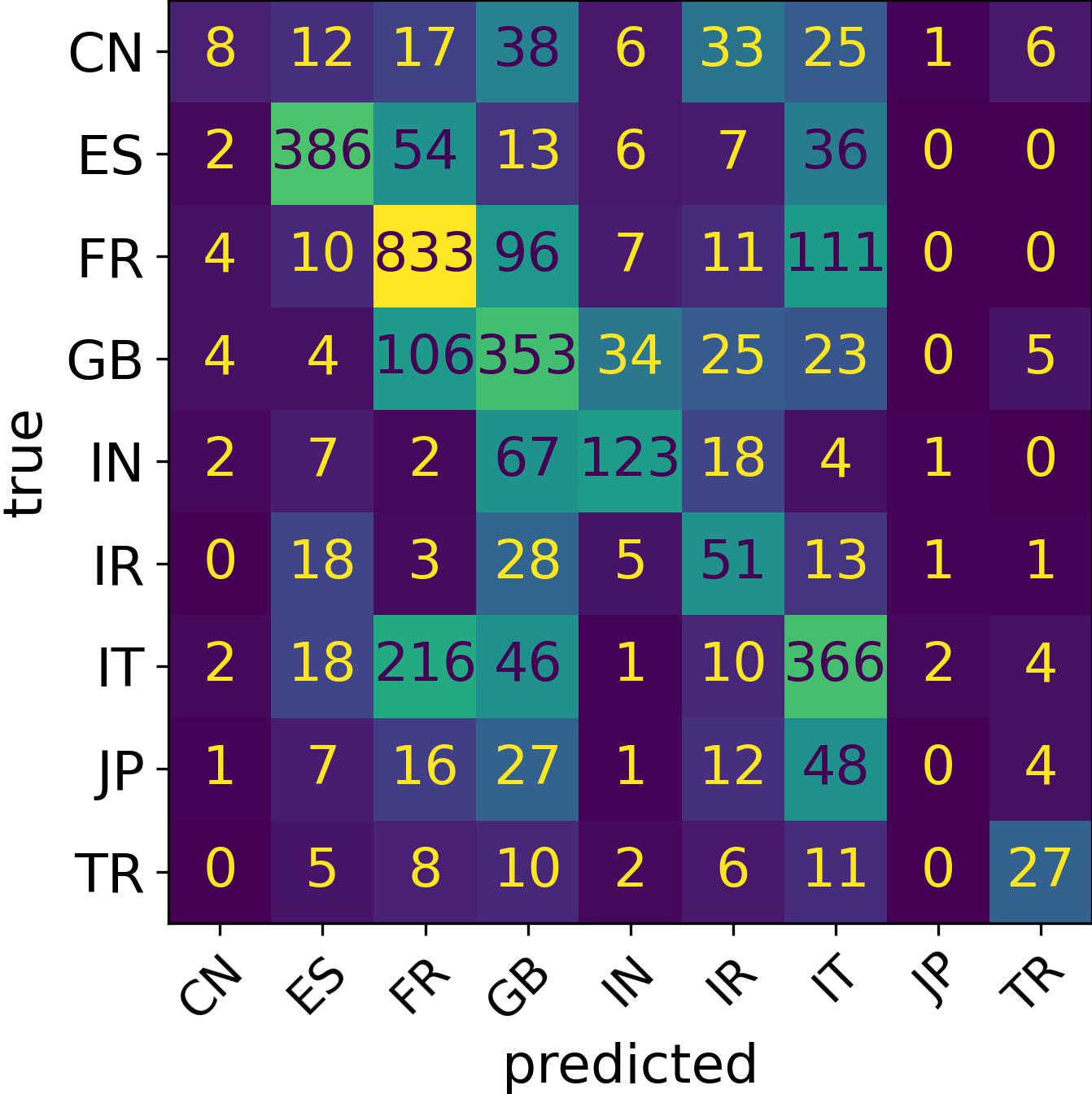}
         \caption{place}
         \label{fig:tab:cm:place}
     \end{subfigure}
          \hfill
     \begin{subfigure}[b]{0.2\textwidth}
         \centering
         \includegraphics[width=\textwidth]{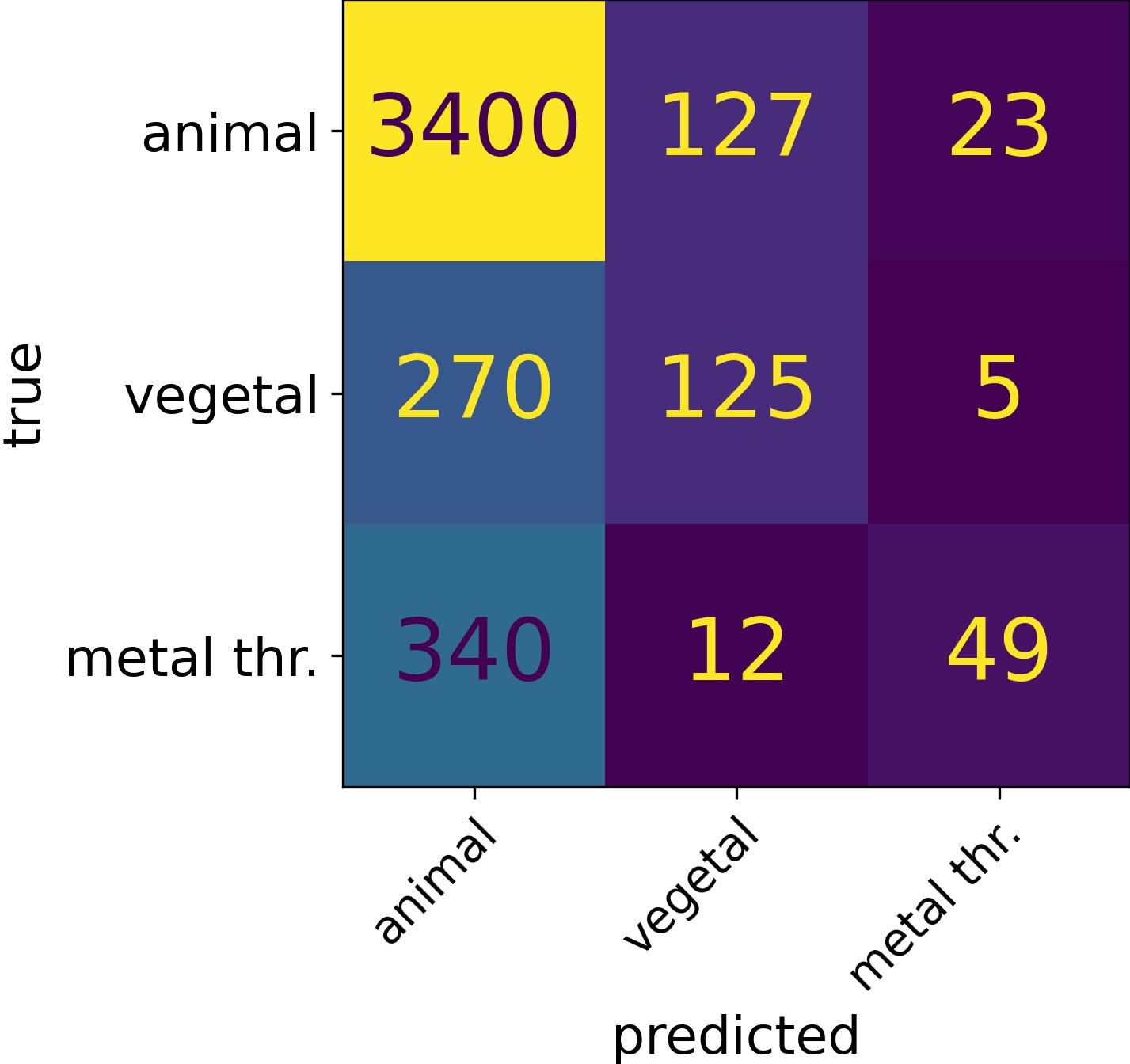}
         \caption{material}
         \label{fig:tab:cm:material}
     \end{subfigure}
     \hfill
     \begin{subfigure}[b]{0.25\textwidth}
         \centering
         \includegraphics[width=\textwidth]{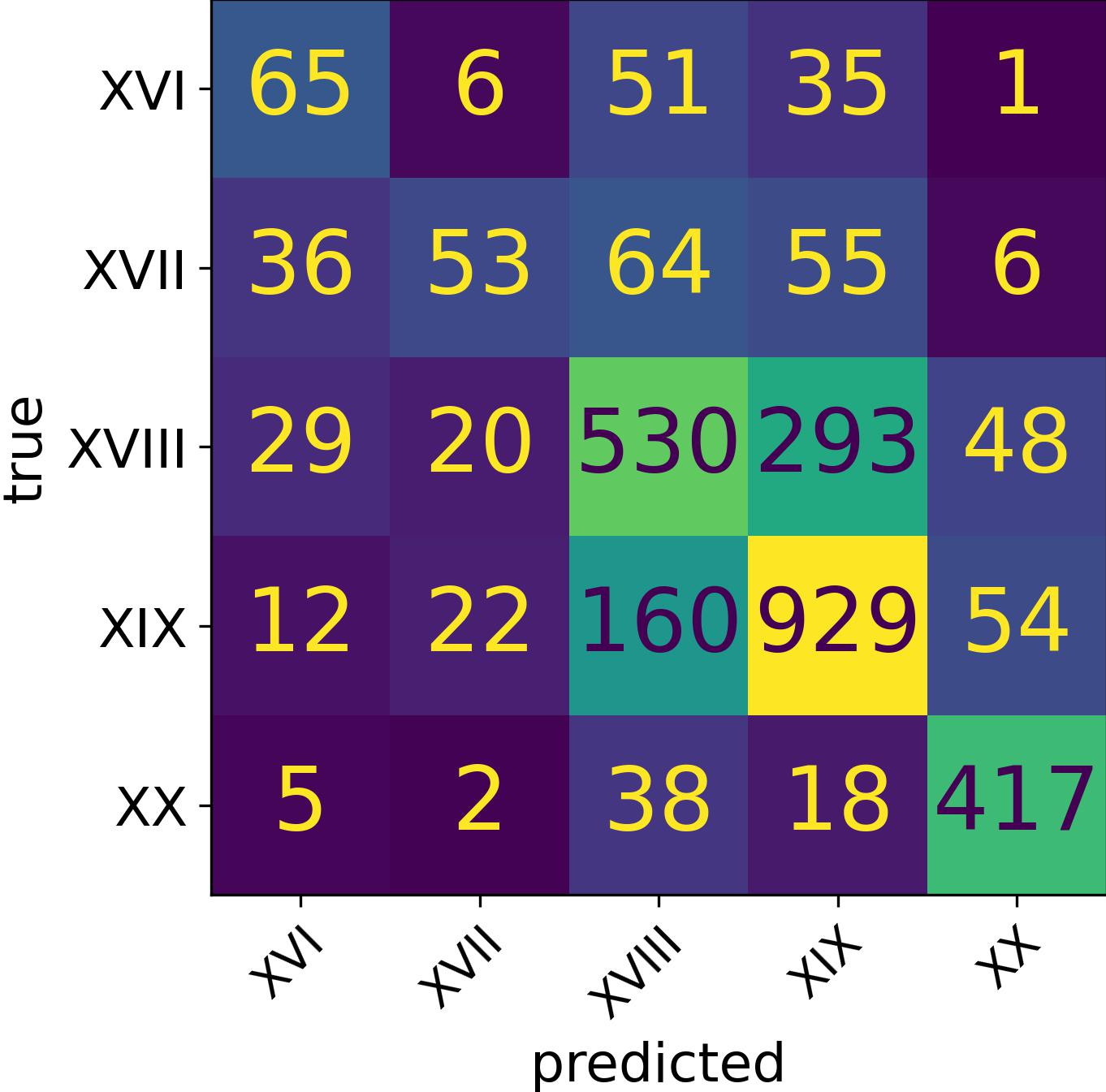}
         \caption{timespan}
         \label{fig:tab:cm:timespan}
     \end{subfigure}
     \hfill
     \begin{subfigure}[b]{0.2\textwidth}
         \centering
         \includegraphics[width=\textwidth]{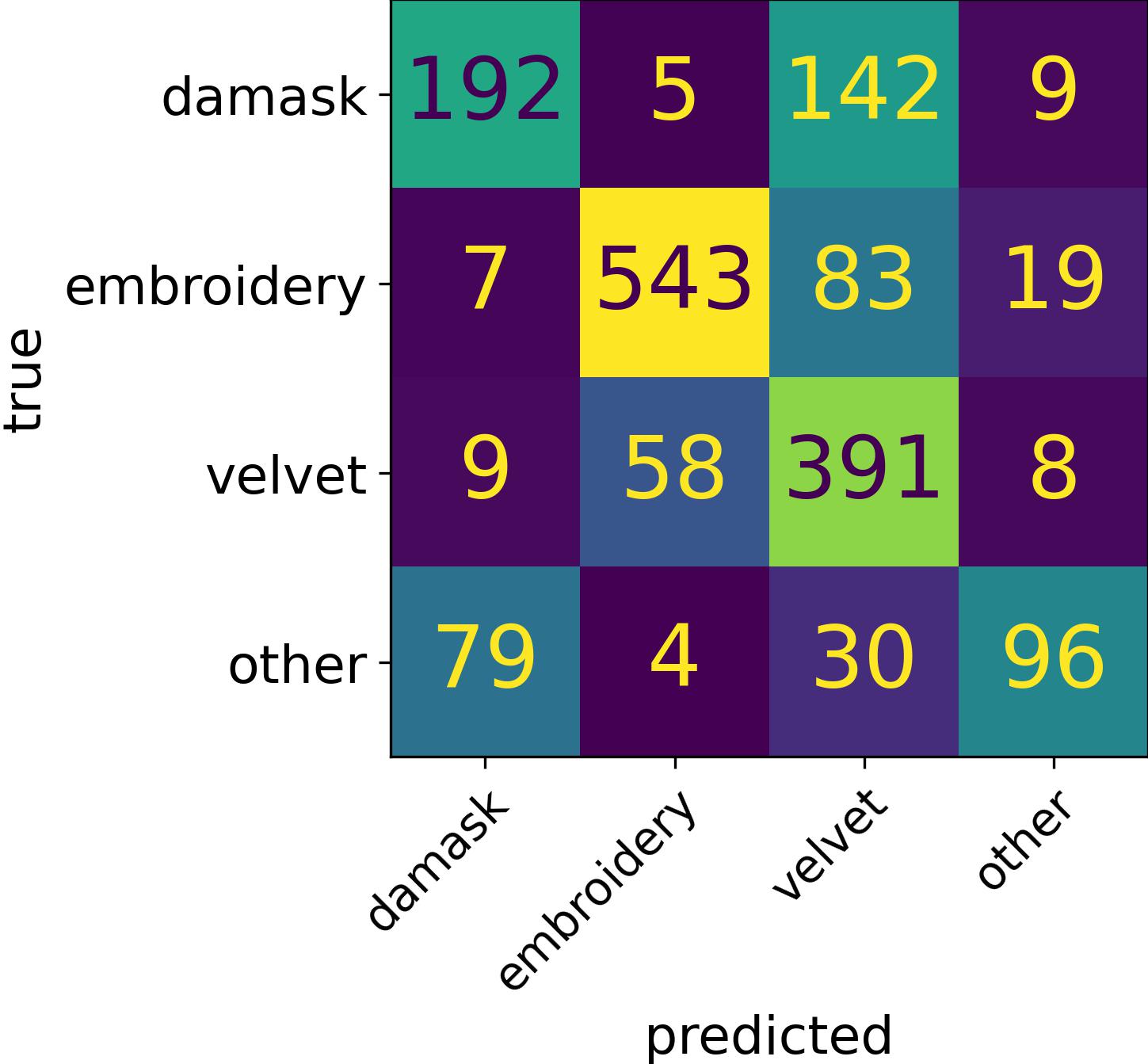}
         \caption{technique}
         \label{fig:tab:cm:technique}
     \end{subfigure}

    \caption{Tabular classifier confusion matrices}
        \label{fig:tab:cm}
\end{figure}

Figure~\ref{fig:tab:cm} shows the confusion matrices for the tabular classifier. Here, the errors seem entirely in line with class imbalance, i.e., the classifier tends to incorrectly predict common classes more often than it incorrectly predicts uncommon classes. This effect seems significantly more pronounced than in other classifiers. We guess that, given the relatively few features and their coarseness, priors end up having more importance. Intuitively, the fact that "museum" is the most important feature (Table~\ref{tab:tab_feature_importance}) for all tasks, might suggest that it is acting as a conditioning variable on the priors.

\subsection{Multimodal classification}
As pointed out in Section~\ref{sec:exp:multimodal}, the comparison of the classification accuracies 
indicates that the text classifier achieves the best performance of all modalities (Table~\ref{tab:mtl_dev_test_eval_text}), but of course, it is only applicable when text is available, which is only the case for a relatively low number of records (cf.~Table~\ref{tab:modality_stats}).

\begin{figure}[h]
     \centering
     \begin{subfigure}[b]{0.25\textwidth}
         \centering
         \includegraphics[width=\textwidth]{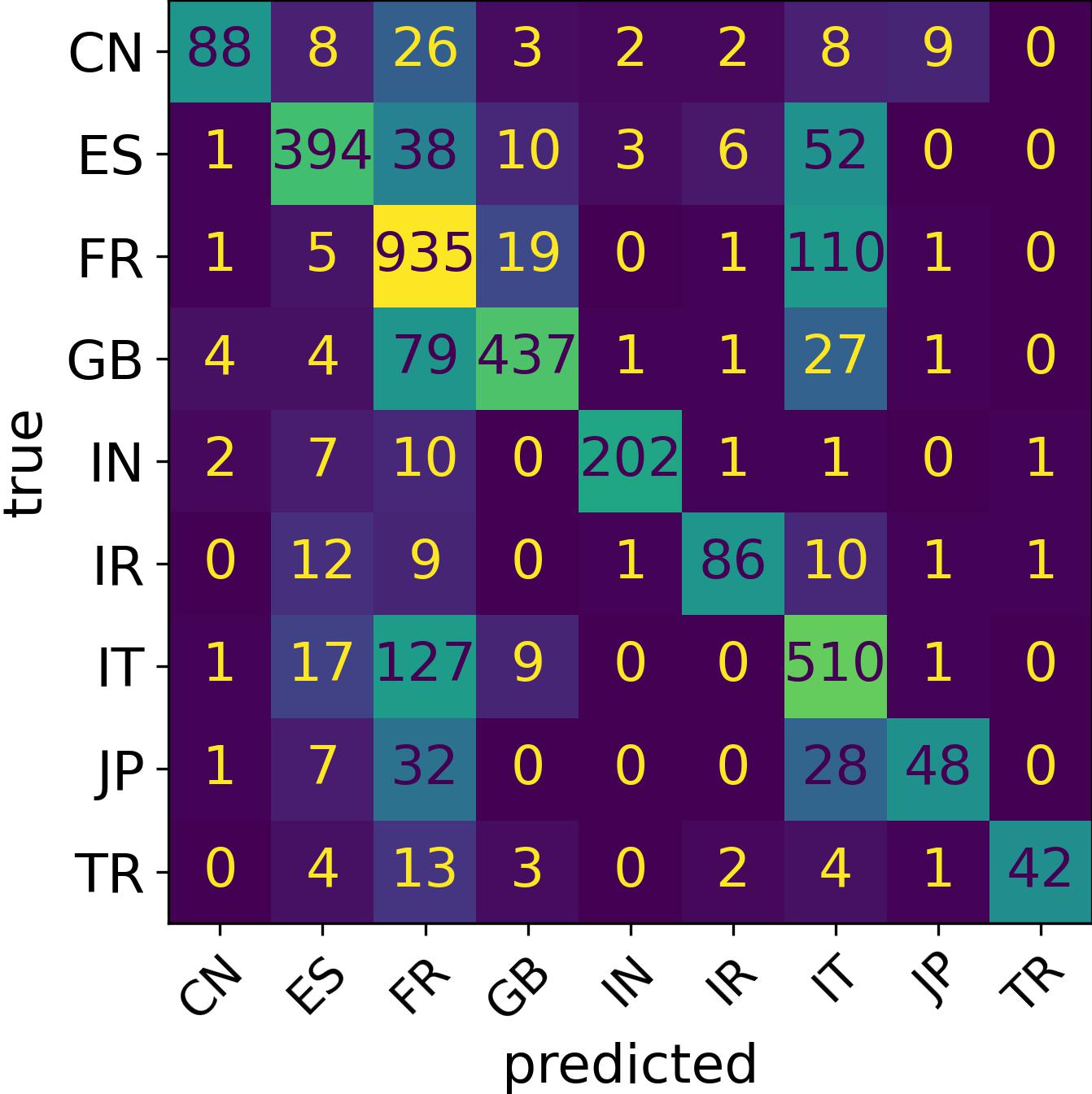}
         \caption{place}
         \label{fig:multimodal:cm:place}
     \end{subfigure}
          \hfill
     \begin{subfigure}[b]{0.20\textwidth}
         \centering
         \includegraphics[width=\textwidth]{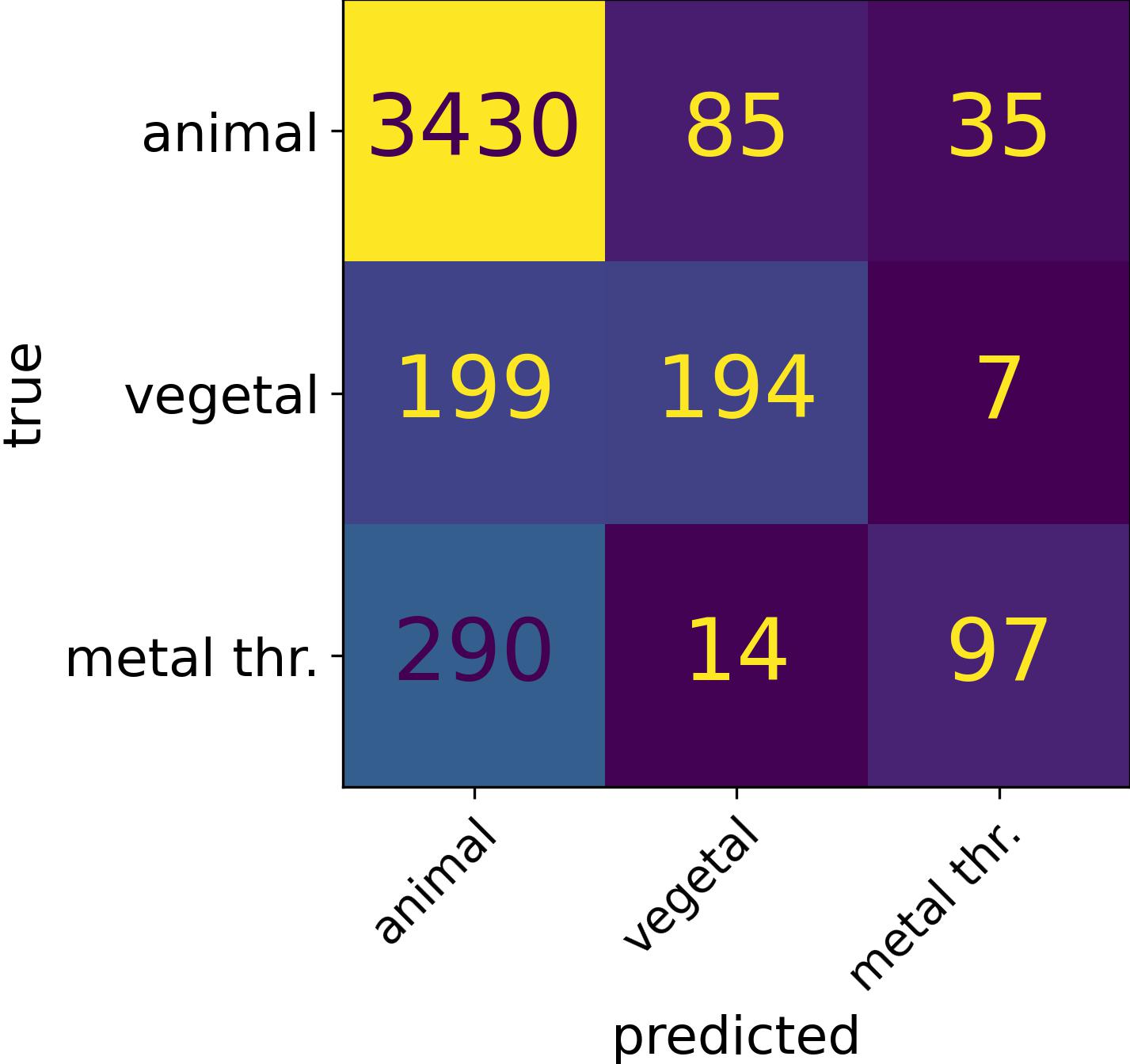}
         \caption{material}
         \label{fig:multimodal:cm:material}
     \end{subfigure}
     \hfill
     \begin{subfigure}[b]{0.25\textwidth}
         \centering
         \includegraphics[width=\textwidth]{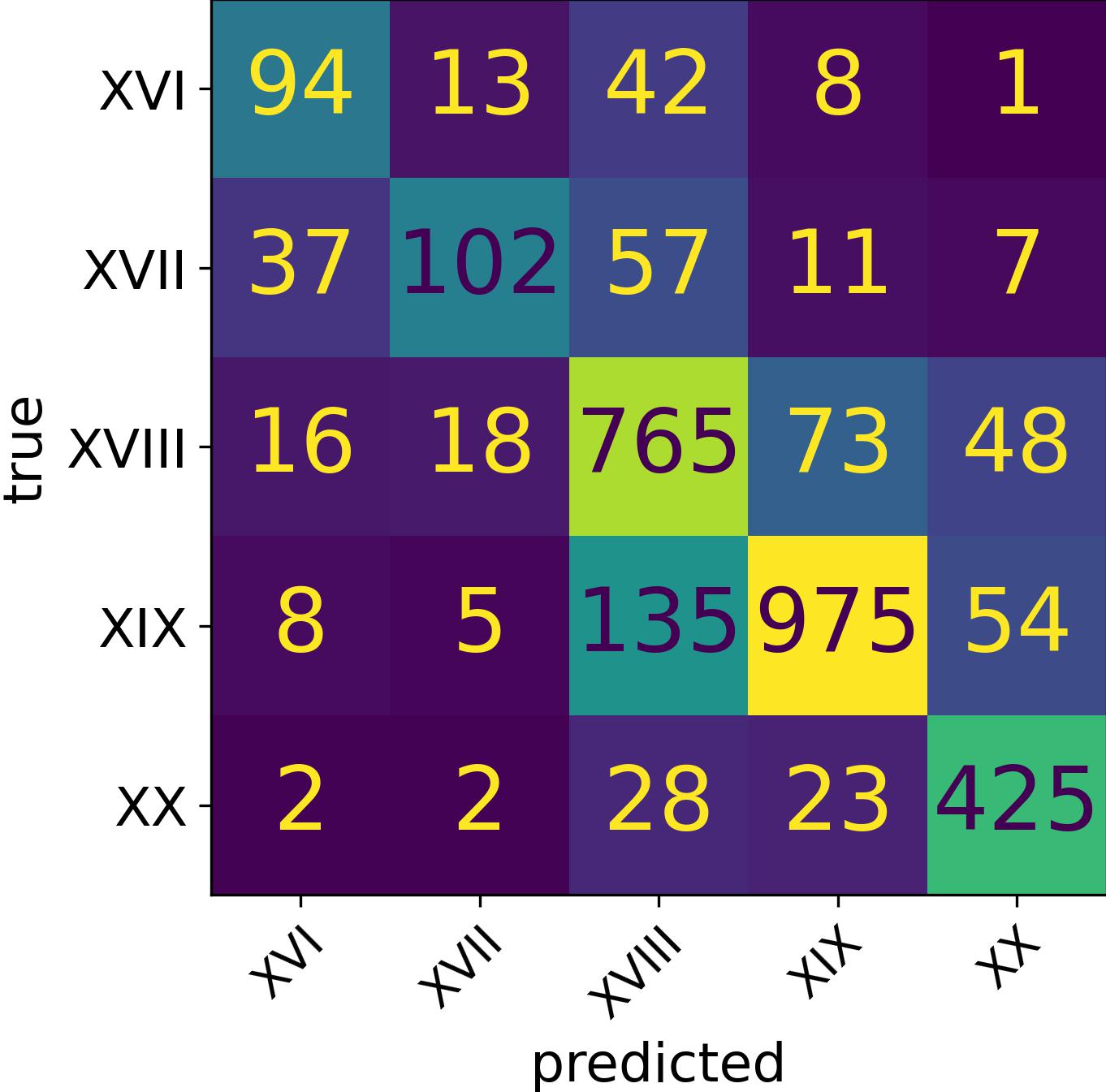}
         \caption{timespan}
         \label{fig:multimodal:cm:timespan}
     \end{subfigure}
     \hfill
     \begin{subfigure}[b]{0.20\textwidth}
         \centering
         \includegraphics[width=\textwidth]{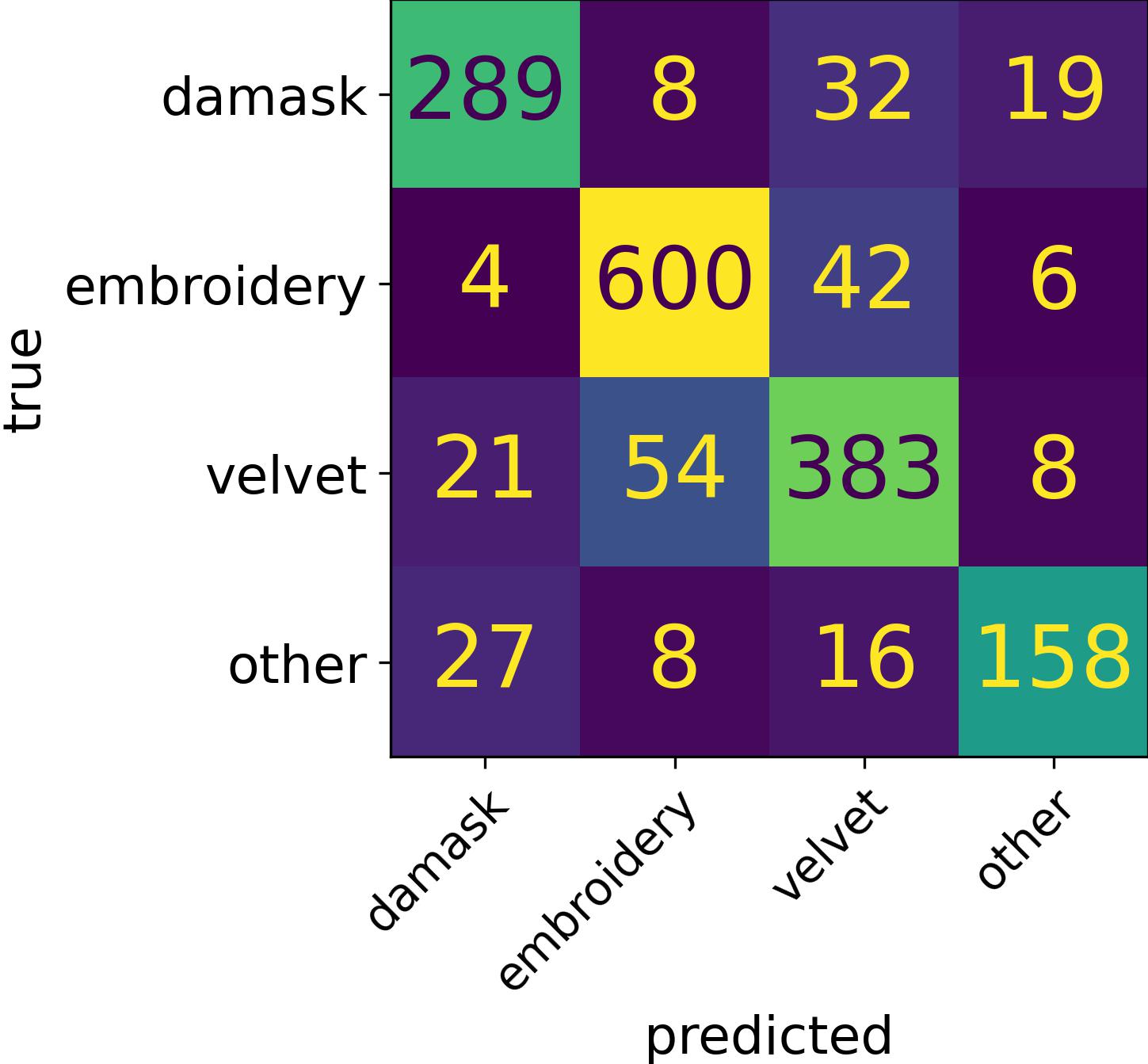}
         \caption{technique}
         \label{fig:multimodal:cm:technique}
     \end{subfigure}

    \caption{Multimodal classifier confusion matrices}
        \label{fig:multimodal:cm}
\end{figure}

This confirms our hypothesis that multimodal classification results in a better classification performance if one of the aspects under consideration is to obtain correct predictions for a number of records that is as large as possible. Figure~\ref{fig:multimodal:cm} shows the confusion matrices for multimodal classification. Most errors in the timespan task occur between chronologically similar dates. Most errors in the place task occur between countries that are geographically close to each other, e.g., Italy (IT) and France (FR). The tendency of errors in these tasks is towards both similar and more common labels (class imbalance). As far as material is concerned, errors occur primarily between animal fiber and the other labels. This is because all objects are made of animal fibre, silk, and possibly due to the much more significant label imbalance among a small set of labels. No clear trend can be observed for the prediction of technique, including no significant trend toward predicting the most common classes. This is perhaps due to the smaller imbalance in this task.

\section{Integrating and visualizing the predictions}
\begin{figure*}[htbp]
    \centering
    \includegraphics[width=0.8\textwidth]{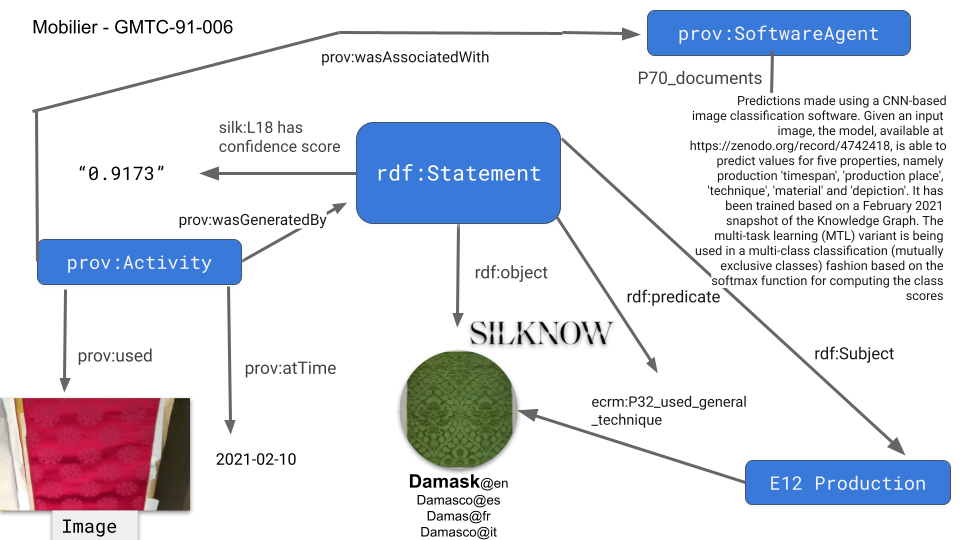}
    \caption{Graph showing the prediction of the production technique (damask) with a high confidence score (0.9173) using the textual analysis software.}
    \label{fig:prediction-scheme}
\end{figure*}

To model the predictions as part of the SILKNOW Knowledge Graph ontology we use classes and properties of the Provenance Data Model (Prov-DM), more specifically the
PROV ontology (PROV-O)\footnote{\url{https://www.w3.org/TR/prov-dm/}}, an OWL2 ontology. It makes it possible to map PROV-DM to RDF. Furthermore, it allows the expression of important elements of the predictions for each modality. These different predictions can be represented using different \texttt{prov:activity} classes each. The image, text description, or categories each prediction is based on are represented with the property \texttt{prov:used}. The exact date of the prediction is represented with prov:atTime and prov:wasAssociatedWith connects the activity class to the \texttt{prov:SoftwareAgent} class, which is used to describe the particular algorithm and model used. The predicted metadata value is represented with rdf:Statement, connected to prov:activity via a \texttt{prov:wasGeneratedBy} property. The confidence score of the prediction is expressed through the property L18 ("has confidence score") from our own SILKNOW Ontology. The predicted value is expressed in form of a URI with \texttt{rdf:object}, the type of the predicted property through rdf:predicate and its fitting CIDOC-CRM property type. The property \texttt{rdf:subject} connects the statement to the production class (E12) of the object in the Knowledge Graph.
Every prediction is inserted in the appropriate part of the existing KG. For example, if a material value gets predicted, it gets inserted with the CIDOC-CRM property \texttt{P126\_employed} at the production class of the object. See Figure~\ref{fig:prediction-scheme} for an illustration of the data model. 

The prediction models were only trained on group labels (Section~\ref{sec:label_groups}), and can thus only predict those. It is sometimes necessary to map them back to a more concrete concept of the SILKNOW Thesaurus. If, for example, "Damask" is predicted in the form of its facet link \url{http://data.silknow.org/vocabulary/facet/damask} it will be automatically converted to \url{http://data.silknow.org/vocabulary/168}, as facet links are too general for concrete category values. All predictions are converted one after another using the described data model and saved in the Terse RDF Triple Language (TTL) file format which is uploaded and stored as its own graph identified by \url{http://data.silknow.org/predictions}. This makes it possible to always identify and eventually separate predictions from original values obtained from the museums. In total, 98,379 predictions were made for 19,248 distinct objects. These were uploaded into the SILKNOW Knowledge Graph.

\begin{figure*}[htbp]
    \centering
    \includegraphics[width=0.5\textwidth]{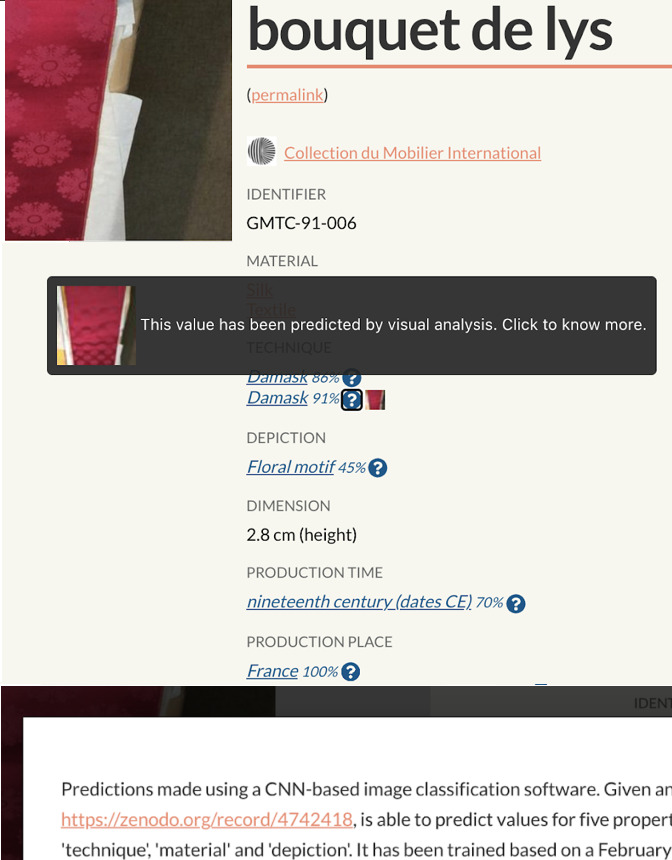}
    \caption{UI Screenshot showing the prediction of the production technique (damask) with a high confidence score (0.9173) using the image analysis software.}
    \label{fig:adasilk-prediction-example}
\end{figure*}

In our exploratory search engine ADASilk, the predictions are displayed differently from values that come originally from the museums: They are shown in blue together with their confidence score as a percentage next to them. A tooltip is available to explain how this value was predicted, including the modality, algorithm, model identifier, etc. In order to display predictions like this on ADASilk, the respective SPARQL query was updated and new subqueries were used to take the aforementioned new properties into account. See Figure~\ref{fig:adasilk-prediction-example} for a screenshot.

\section{Conclusions}\label{sec:conclusions}
We presented results for three individual modality classifiers, as well as multimodal results. In terms of our original hypothesis, presented in Section~\ref{sec:introduction-hypothesis}, we showed that we were in fact able to accurately predict missing properties in the digitized silk fabric artifacts that made up our dataset. While the quality of predictions varied between individual modalities, we showed that the multimodal approach provided by far the best results (74.2\% vs 55.6\%) and even that combining any two modalities outperforms any single modality.

To recapitulate, our contributions included the already mentioned multimodal approach tailored specifically to the challenges we faced in the multimodal scenario, including the incomplete overlap of data across modalities. The individual approaches of each modality-specific classifier also provide a useful contribution to the automated classification of cultural heritage objects. The image and text classifier both offer the possibility of being applied to data outside a Knowledge Graph (KG) or database, possibly even directly submitted by a system user. The tabular classifier, on the other hand, offers the possibility of classifying data in a KG or database when no text descriptions or images are present by relying on other properties. The text classifier provides the best results of any single modality when text is present. Including text descriptions when classifying cultural heritage objects is thus highly advantageous and should be considered in any practical work. It is also important to remember that in most practical situations, including inside a KG or other knowledge bases, images are more common than text descriptions of objects in the cultural heritage domain, and thus image classification may be the only viable approach to automatically assigning metadata to a large portion of any cultural heritage dataset. When adjusted for records with missing modalities, because images are more common in records than text descriptions, the image classifier actually outperforms the text classifier. Although useful in fewer circumstances, once also adjusted for missing modalities, the tabular classifier is the best-performing single modality classifier. This modality is probably the best starting point for predicting missing metadata inside a KG or a database.

For both text and image classification, significant improvement of results will probably have to lean more heavily into self-supervised learning, as supervised data scarcity is likely to remain a challenge in the near future. This applies to individual modalities and to the fusion approach. Similarly, class imbalance and label noise are challenges for all modalities, and different ways to handle them could be found not just at the level of the individual modalities, but perhaps also with a change of fusion approach.

\subsection{Data}
The data we used in our work originally comes from many museum sources and is from a very specific Cultural Heritage domain: historical, European silk fabrics. We applied common methods to process such data and developed an ontology and a Knowledge Graph out of the original museum texts and images. Such an effort comes typically with challenges, which in our case consisted mostly of a small amount of (training) data, domain specificity, different styles of writing texts and capturing images of objects, different languages (in the case of texts), and finally simply annotation errors, typos, and other errors that happened during the original digitization. Not all of these challenges could be completely overcome. In fact, the metadata gaps constitute part of the motivation to conduct this research work. The exclusion of some data from our dataset was necessary to ensure that the classes present had a viable number of samples. This was, however, very much alleviated through the grouping process, which was possible through our domain expert-designed thesaurus about silk fabric concepts. In the end, we were able to publish a cultural heritage dataset that can be used for automated classification for multiple modalities and enable the study of multimodal approaches. In this work, we also provide the data modeling of how metadata predictions for data such as ours can be represented within knowledge graphs or other knowledge bases.

\subsection{Image classification}
We have shown that the properties of silk fabrics can be predicted from images of these fabrics. 
In this context, we proposed to use the focal loss for training in order to compensate for the effects of class imbalance in the training set, a problem that is quite common in the cultural heritage domain. 
Our results indicate that the proposed strategy can mitigate this problem to a certain degree, in particular by improving the classification performance for the underrepresented classes in terms of the F1 score. 
Image classification performs particularly well for the task of predicting the technique used for producing a fabric. 
Nevertheless, there is still room for improvement, as indicated by the performance metrics for all variables. 
Future work on image classification could concentrate on improving the performance of underrepresented classes even more, e.g., by using methods for few-shot learning. Furthermore, as some experimental results indicated that some training labels might be incorrect, training methods that are robust against such errors ("label noise") could be investigated.

\subsection{Text classification}
When text descriptions are present, the text classifier provides the best results of any single modality. It seems, thus, that the text classifier was able to overcome the primary challenges it faced: small dataset, domain specificity, cross-linguality, and museum-specific text styles. This was primarily achieved by the choice of XLM-R as the basis of the text classifier. 
Our results indicate that the use of the focal loss, originally introduced in computer vision, can help improve results in text classification and mitigate class imbalance within the cultural heritage domain. Future work can include the evaluation of more recent advances in loss functions for class imbalance. While more focus on class imbalance is desirable, future work must also deal with noisy labels, as this is an issue in the dataset and is very likely to be an issue in any practical application within this domain. 
Misleading text descriptions stand out as a challenge for text classification that might be very difficult to overcome without a significant increase in data quantity for silk heritage or leveraging training from the broader cultural heritage domain.

\subsection{Tabular classification}
Tabular classification was not emphasized in this work due to its more limited uses. Despite this, it proved very useful and provided surprisingly good results when considering that, by definition, this modality is present in database records. In the future, class balance could be tackled by generating artificial examples using nearest neighbor data generation techniques common in tabular classification. The inputs to the algorithm could also be switched to finer-grained labels (e.g., "cotton" instead of "vegetable fibre") as the requirements for the number of examples are more relaxed in inputs than in target labels. Other methods could also be explored. Neural Networks have become competitive algorithms, in certain cases, for tabular classification. It would be interesting, in this case, to replace the one-hot input labels with word embeddings that could make clear the similarities between them (e.g., "cotton" and "linen" are both vegetable fibres).

\subsection{Multimodal classification}
When all data is considered, we have shown that the multimodal approach is the best according to the macro-averaged F1 metric. While most records contain images, not all do (3.4\%), and a smaller number of records contain neither text nor images (2.1\%). On the other hand, if we had tried to implement a classifier using the text modality alone, we could only classify 40\% of the records. While we can say that a multimodal approach does allow us to classify a greater number of records than using images alone, the primary practical benefit of the multimodal approach over performing just image classification is probably the qualitative improvement in classification results demonstrated. 
In terms of future work for the fusion classifier, an easy-to-implement option could be to add softmax scores as inputs, although adding the entire vector for each modality might run up against a dimensionality challenge as the number of examples used to train the fusion classifier is relatively small. A comprehensive analysis of this approach was beyond the scope of this work. Perhaps different approaches to integrating predictions from multiple modalities beyond fusion could also prove fruitful. For example, using the predictions of image and text modal to replace missing non-target values in the input to the tabular classifier. 

\section*{Acknowledgments}
This work was supported by the Slovenian Research Agency and the European Union's Horizon 2020 research and innovation program under SILKNOW grant agreement No. 769504.

\section*{Data Availability}
The dataset used in this article is publicly available at \url{https://zenodo.org/record/6590957}. The entire Knowledge Graph from which the dataset was created is also available at \url{https://zenodo.org/record/5743090}. Finally, the thesaurus used in this work is available at \url{https://skosmos.silknow.org/thesaurus/}.

\section*{Code Availability}
The code used to perform the experiments reported in this work, namely all the classifiers, is available online at \url{https://github.com/silknow/multimodal_cultural_heritage}. The tools used to create the dataset are also publicly available. The crawler is available at \url{https://github.com/silknow/crawler/} and the converter at \url{https://github.com/silknow/converter/}. Other code relevant or used is this work may be found in the project collection of code repositories at \url{https://github.com/silknow/}.

\bibliographystyle{unsrt}
\bibliography{biblio}

\end{document}